\documentclass{article} 
\usepackage{iclr2026_conference,times}

\usepackage{amsmath,amsfonts,bm}

\def\figref#1{figure~\ref{#1}}

\def\secref#1{section~\ref{#1}}

\def\eqref#1{equation~\ref{#1}}

\def\1{\bm{1}}

\DeclareMathAlphabet{\mathsfit}{\encodingdefault}{\sfdefault}{m}{sl}
\SetMathAlphabet{\mathsfit}{bold}{\encodingdefault}{\sfdefault}{bx}{n}

\usepackage{hyperref}
\usepackage{url}
\usepackage{graphicx}
\usepackage{multirow}
\usepackage{booktabs}
\usepackage{xspace}
\usepackage{enumitem}
\usepackage{wrapfig}
\usepackage{xcolor}
\usepackage{subcaption} 

\newtheorem{lemma}{Lemma}
\newtheorem{proof}{Proof}

\newtheorem{theorem}{Theorem}

\newcommand\tabref[1]{Tab.~\ref{#1}}
\renewcommand\secref[1]{Sec.~\ref{#1}}
\renewcommand\figref[1]{Fig.~\ref{#1}}

\newcommand{\fakeparagraph}[1]{\vspace{1mm}\noindent\textbf{#1.}}
\newcommand{\sysname}{PhaseFormer\xspace}

\newcommand{\red}[1]{\textcolor{red}{#1}}
\newcommand{\blue}[1]{\textcolor{blue}{#1}}

\newcommand{\best}[1]{\red{\textbf{#1}}}
\newcommand{\second}[1]{\blue{\underline{#1}}}

\ifodd 0

\else

\fi

\title{PhaseFormer: From Patches to Phases for Efficient and Effective Time Series Forecasting}

\author{
Yiming Niu\thanks{Equal contribution.} \\
School of Computer Science and Engineering \\
Beihang University \\
Beijing, China \\
\texttt{yimingniu@buaa.edu.cn}
\And
Jinliang Deng\footnotemark[1] \\
Department of Computer Science and Engineering \\
The Hong Kong University of Science and Technology \\
Hong Kong SAR, China \\
\texttt{dengjinliang@ust.hk}
\And
Yongxin Tong\thanks{Corresponding author.} \\
School of Computer Science and Engineering \\
Beihang University \\
Beijing, China \\
\texttt{yxtong@buaa.edu.cn}
}

\iclrfinalcopy 
\begin{document}

\maketitle

\begin{abstract}
Periodicity is a fundamental characteristic of time series data and has long played a central role in forecasting. Recent deep learning methods strengthen the exploitation of periodicity by treating patches as basic tokens, thereby improving predictive effectiveness. However, their efficiency remains a bottleneck due to large parameter counts and heavy computational costs. This paper provides, for the first time, a clear explanation of why patch-level processing is inherently inefficient, supported by strong evidence from real-world data. To address these limitations, we introduce a phase perspective for modeling periodicity and present an efficient yet effective solution, PhaseFormer. 
PhaseFormer features phase-wise prediction through compact phase embeddings and efficient cross-phase interaction enabled by a lightweight routing mechanism. Extensive experiments demonstrate that PhaseFormer achieves state-of-the-art performance with around 1k parameters, consistently across benchmark datasets. Notably, it excels on large-scale and complex datasets, where models with comparable efficiency often struggle. This work marks a significant step toward truly efficient and effective time series forecasting.
Code is available at this repository: \url{https://github.com/neumyor/PhaseFormer_TSL}.
\end{abstract}

\section{Introduction}
Time series forecasting underpins decision-making across diverse domains such as finance, energy, climate science, and healthcare, playing a pivotal role in tasks including weather forecasting~\citep{weathernature2025,wu2021autoformer}, energy consumption planning~\citep{lai2018energy,alvarez2010energy,cheng2021windturbine}, traffic scheduling~\citep{cirstea2022towardstraffic,cirstea2021enhancenet,wu2021autocts}.
In recent years, deep learning has demonstrated promising potential in this field by leveraging end-to-end modeling and powerful representational capacity to extrapolate from history to future trends. 

A central inductive bias in forecasting models is periodicity--the recurring temporal structure inherent in many real-world time series. 
Recent advances exploited this property by segmenting sequences into patch tokens, potentially aligned with cycles, prior to processing by the crafted models~\citep{nie2023patchtst, zhang2023crossformer, huang2025timebase, tang2025unlocking}. 
For instance, \cite{nie2023patchtst} applied Transformer to tokenized time series to capture temporal correlations within and between cycles, while \cite{zhang2023crossformer} extended this paradigm by modeling cross-dimension dependencies and cross-scale interactions.

\begin{figure}[t]
  \centering
  \begin{subfigure}[t]{0.49\textwidth}
    \centering
    \includegraphics[width=\linewidth]{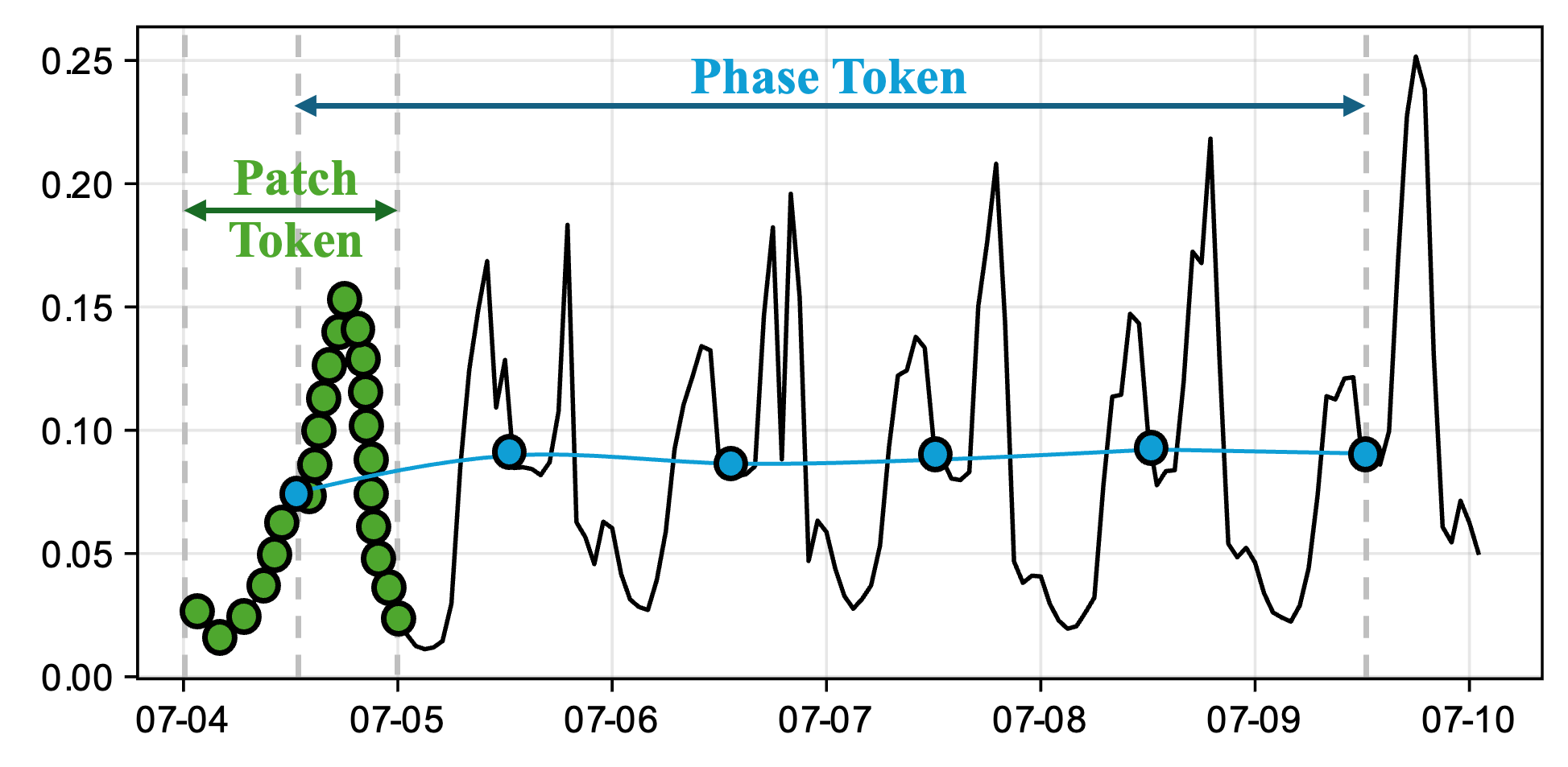}
    \caption{Patch Token vs. Phase Token.}
    \label{fig:phase_token}
  \end{subfigure}
  \hfill
\begin{subfigure}[t]{0.49\textwidth}
    \centering
    \includegraphics[width=\linewidth]{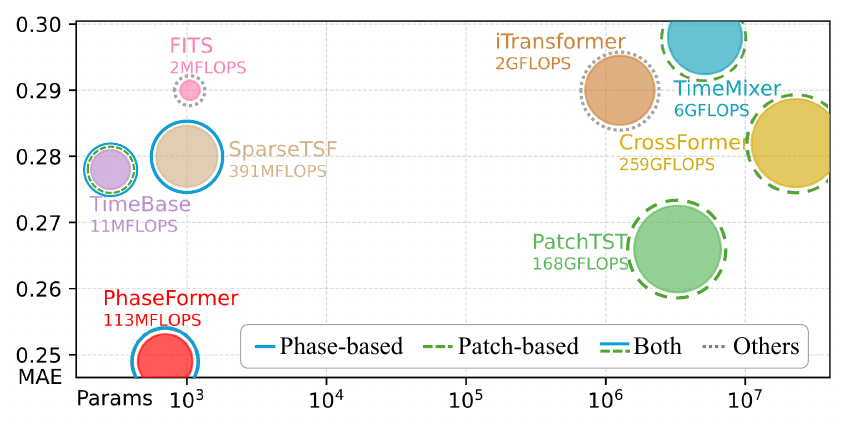}
    \caption{Model Accuracy and Efficiency Comparison.}
    \label{fig:efficiency_plot}
  \end{subfigure}
  \caption{Comparison between patch-based and phase-based representations for time-series forecasting. (a) illustrates the difference in tokenization. (b) jointly evaluates model accuracy, parameter scale, and computational overhead on the Traffic dataset, where marker size indicates FLOPS.}
  \label{fig:introduction}
\end{figure}

Despite their effectiveness, patch-based approaches struggle to scale efficiently to large and complex datasets~\citep{nie2023patchtst, zhang2023crossformer,tang2025unlocking}. 
\emph{We attribute this poor scalability to the substantial variability of cycle patterns in real-world scenarios}. This variability stems from dynamic external factors, which continuously shift the cycle patterns. For instance, traffic flow patterns may evolve as new infrastructure is introduced, while electricity demand can change with adjustments in work schedules. This variability forces models to construct a high-dimensional representation space to faithfully accommodate the broadened distribution, which inevitably inflates both parameter counts and computational costs~\citep{nie2023patchtst, zhang2023crossformer}. Additionally, these methods also struggle to generalize under such varying behavior, resulting in unreliable forecasts for samples beyond training data.

To address this challenge, we introduce a novel phase-based perspective that focuses on values aligned at the same offset across successive cycles. From this perspective, the dynamics of a time series are characterized by the cross-period trends of each phase--captured as phase tokens--while disregarding the full cyclic behavior. As illustrated in Fig.~\ref{fig:phase_token}, phase tokens exhibit significantly lower variability than patch tokens, enabling more efficient and generalizable representation. Importantly, excluding cycle patterns has minimal impact on forecasting effectiveness, since the cyclic behaviors remain locally stable and thus require little effort to predict. We study and verify these properties in depth in Sec.~\ref{sec:motivations} using real-world data, showing the stationarity and compactness of the feature space offered by phase tokenization. 

Building on these insights, we propose Phase-based Routing Transformer, abbreviated as \textbf{\sysname}, which reframes time series as a collection of phase tokens and casts step-wise prediction as phase-wise prediction. Specifically, \sysname (i) aligns and extracts phase tokens from the input sequence and maps them into a shared low-dimensional latent space, (ii) employs a lightweight routing mechanism to enable efficient communication across phases, and (iii) applies a shared predictor to project the latent representations into forecasts for each phase. Extensive experiments demonstrate that, compared with PatchTST~\citep{nie2023patchtst} and Crossformer~\citep{zhang2023crossformer}, \sysname achieves over \textbf{99.9\%} reduction in both parameter count and computational cost, while delivering consistent improvements in prediction accuracy across all seven benchmark datasets, as illustrated by the Traffic dataset in Fig.~\ref{fig:efficiency_plot}. 
Moreover, in contrast to methods with comparable efficiency  such as SparseTSF~\citep{lin2024sparsetsf} and TimeBase~\citep{huang2025timebase}, \sysname significantly enhances predictive effectiveness, particularly on large and complex datasets. 
Finally, we conduct a comprehensive analysis of different configurations to reveal the necessity of the constructed components and the effects of various hyperparameters. Our contributions are as follows:

\begin{enumerate}[leftmargin=2em]
\item We introduce a phase-based perspective that aligns values across cycles for the characterization of long-term time series, empirically and theoretically demonstrating improved feature stationarity and compactness over the patch-based perspective.

\item We propose \sysname, a lightweight forecasting model that reframes time series as phase tokens, maps them into a shared latent space, and employs a routing mechanism with a shared predictor to enable efficient phase-wise forecasting.

\item Extensive experiments are conducted to showcase that \sysname achieves substantial efficiency gains while consistently improving forecasting accuracy, establishing a superior efficiency–effectiveness trade-off across diverse benchmarks.
\end{enumerate}

\section{Related Works}

\fakeparagraph{Transformer-Based Forecasting Architectures}
Early Transformer-based models for long sequence forecasting often overlooked the periodicity in time series~\citep{zhou2021informer, logtrans2019}. 
Subsequent research introduced domain-specific priors that better understand recurring temporal structures. 
Autoformer~\citep{wu2021autoformer} and FEDformer~\citep{zhou2022fedformer} incorporated decomposition strategies and frequency-domain modeling enabling explicit representation of seasonal–trend patterns. 
Pyraformer~\citep{liu2021pyraformer} and Crossformer~\citep{zhang2023crossformer} further enriched temporal modeling by embedding multi-scale hierarchies and cross-variable dependencies, while \cite{nonstationarytrans2022} explicitly accounted for distributional shifts. 
More recently, PatchTST~\citep{nie2023patchtst} reframed time series as patch sequences to enable more accurate characterization of sequence-level semantics, followed by an extension to jointly consider spatial and temporal correlations~\citep{ctpatchtst2025}. Generally speaking, these models embed progressively stronger temporal biases, though often at the cost of massive parameter counts and heavy computation.

\fakeparagraph{Efficiency-Oriented Forecasting Models}
A growing body of research emphasizes efficiency, aiming to design lightweight forecasting architectures. 
Patch-based MLP variants such as xPatch~\citep{xpatch2025},TimeMixer~\citep{timemixer24}, and PITS~\citep{PITS2024} exploited compact tokenization or hierarchical dependencies to reduce parameter counts while maintaining accuracy. 
Beyond patches-based methods, frequency-based counterparts leverage spectral representations for compression and denoising.
FreTS~\citep{FreTS2023} applied MLPs in the frequency domain, \cite{filternet2024} learned frequency filters to improve noise robustness, and FITS~\citep{fits2024} achieved strong accuracy with only 10k parameters. 
\cite{deng2024parsimony} demonstrated that selective decomposition can deliver both parsimony and capability.
More recently, SparseTSF~\citep{lin2024sparsetsf} and TimeBase~\citep{huang2025timebase} highlighted the importance of cross-period correlation, sharing a similar motivation with ours.
Despite their impressive computational efficiency, these methods still fall short on forecasting accuracy for large and complex datasets such as Traffic and Electricity~\citep{lin2024sparsetsf, huang2025timebase}. 
Moreover, they lack systematic analysis to answer the fundamental question: \emph{Why can phase tokens serve as an efficient alternative to patch tokens?}
\section{Motivations}
\label{sec:motivations}


\begin{figure}[t]
  \centering
  \begin{subfigure}[t]{0.59\textwidth}
    \centering
    \includegraphics[width=\linewidth]{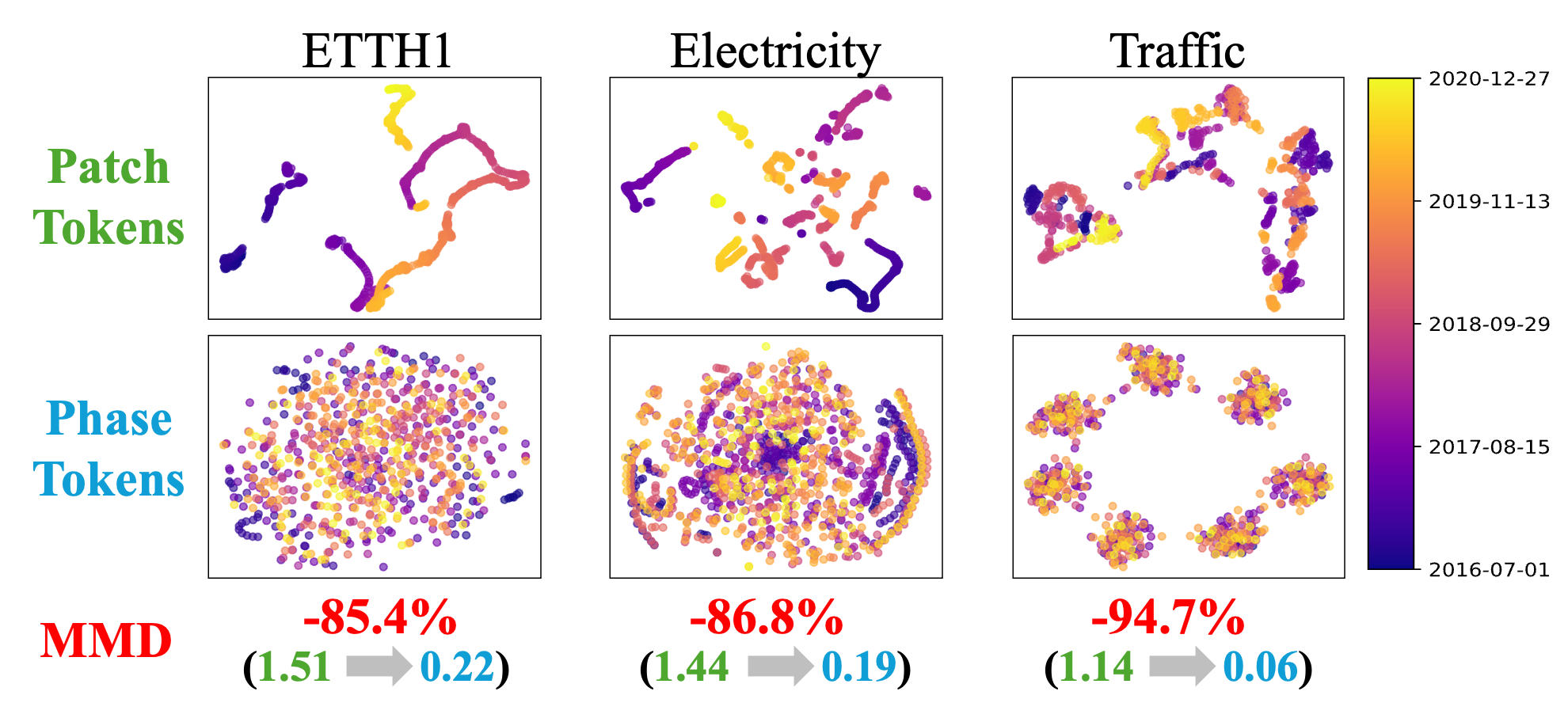}
    \caption{Temporal shift comparison across multiple datasets.}
    \label{fig:visualization_tsne}
  \end{subfigure}
  \hfill
  \begin{subfigure}[t]{0.37\textwidth}
    \centering
    \includegraphics[width=\linewidth]{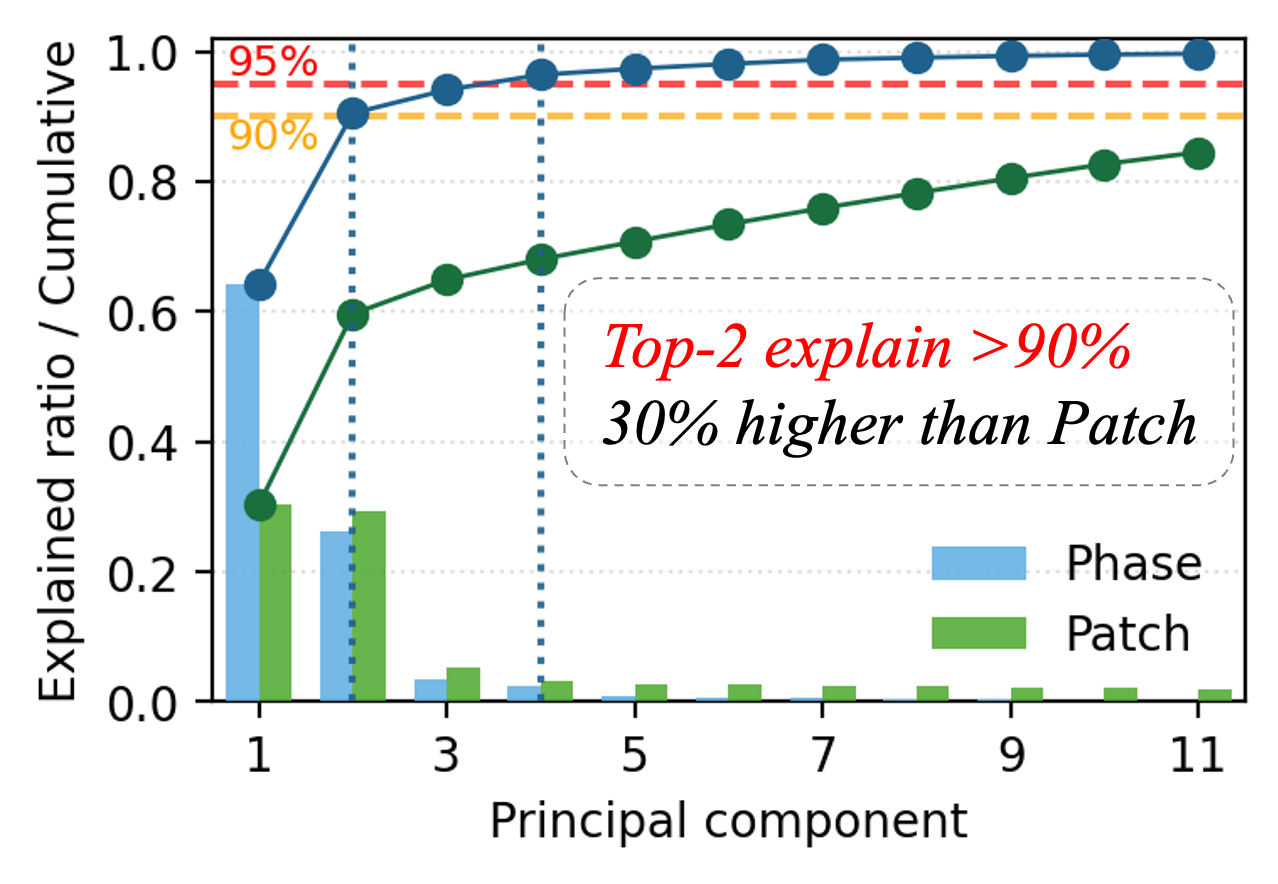}
    \caption{PCA for phase tokens on \emph{Traffic}.}
    \label{fig:traffic_pca_cycle_phase}
  \end{subfigure}
  \caption{Visualization of phase tokenization and its advantages.  
  (a) Phase tokenization yields more stable representations than patch-based embeddings.
  (b) Phase tokens exhibit clear low-dimensionality compared with patch tokens.}  
  \label{fig:phase_token_comparison}
\end{figure}

To motivate our approach, we conduct a comparative analysis of the geometric structures of patch and phase tokens across three widely used datasets. 
As illustrated in \figref{fig:phase_token}, a patch token is composed of adjacent observations within a local period, whereas a phase token is constructed by extracting values at identical offsets across consecutive periods. We gain the following two important insights from the thorough analysis.

\fakeparagraph{Insight 1: Phase tokens are globally stationary, while patch tokens are locally stationary}
To provide an intuitive overview of their geometric structures, we project both types of tokens into two-dimensional spaces using t-SNE~\citep{tsne2008}.
As shown in Fig.~\ref{fig:visualization_tsne}, the distributions of patch tokens drift continuously over time but exhibit local coherence, indicating \emph{local stationarity} and supporting the minimal impact of excluding cycle patterns from intensive processing. In contrast, phase tokens form compact and coherent clusters that remain stable over the long term, reflecting strong \emph{global stationarity}. To rigorously quantify the long-term drift, we compute the average discrepancy distance between each subsequent week and the initial week. Specifically, we adopt the Maximum Mean Discrepancy (MMD) metric~\citep{ouyang2021mmd}, a statistical measure of distributional divergence:
\begin{equation}
\mathrm{MMD}^2(P, Q) 
= \mathbb{E}_{x,x' \sim P}[k(x,x')] 
+ \mathbb{E}_{y,y' \sim Q}[k(y,y')] 
- 2\,\mathbb{E}_{x \sim P, y \sim Q}[k(x,y)],
\end{equation}
where \(P\) and \(Q\) denote tokens collected from two different weeks, respectively, and \(k(\cdot,\cdot)\) is the RBF kernel function. As two distributions become closer, their MMD value approaches zero. The results at the bottom of Fig.~\ref{fig:visualization_tsne} show that the average MMD distance of the phase token space is significantly smaller than that of the patch token space. Taken together, both qualitative and quantitative analyses demonstrate that phase tokenization exhibits substantially lower temporal distribution divergence, thereby \emph{facilitating better generalization across the time axis}.

\fakeparagraph{Insight 2: Phase tokens reside in a lower-dimensional subspace than patch tokens}
To measure the effective dimensionality of the token space, we perform principal component analysis (PCA) on it. Surprisingly, as illustrated in Fig.~\ref{fig:traffic_pca_cycle_phase}, two dimensions are already sufficient to explain over 90\% of the variance of phase tokens, whereas patch tokens require more than eleven dimensions to achieve the same degree of explanation, owing to their drifting behavior observed in Fig.~\ref{fig:visualization_tsne}. Consequently, phase information resides in a low-dimensional subspace, \emph{providing a principled basis for parameter- and computation-efficient modeling}.

We further establish, based on perturbation theory, that phase tokenization remains stable under perturbations of cycle patterns, whereas patch tokenization undergoes structural shifts. Due to space limitations, only the core theorem is presented here, while the detailed proof is provided in \secref{sub:theoretical}.
\begin{theorem}[Phase Tokenization Stability]
Let $X = A G^\top + N \in \mathbb{R}^{D\times H}$ with 
$\operatorname{rank}(A)=\operatorname{rank}(G)=r \ll \min(D,H)$, 
and consider the transformed data
\begin{equation}
X' = X S^\top + R,
\end{equation}
where $\|N'\|_2 \le \|S\|_2\|N\|_2$, 
$\|R\|_2 \le \varepsilon(\|M\|_F+\|N\|_F)$, 
and let $\delta_{\min} > 0$ denote the minimal spectral separation. 
Then there exists a universal constant $C>0$ such that:
\begin{enumerate}[leftmargin=2em]
\item For phase tokenization and corresponding subspace $\mathcal U_r$, there exists:
\begin{equation}
d\big(\mathcal U_r(X),\mathcal U_r(X')\big)
\;\le\;
C\,\frac{\|N\|_2+\|N'\|_2+\|R\|_2}{\delta_{\min}},
\end{equation}
with exact invariance in the noiseless case $(N=R=0)$.
\item For patch tokenization and corresponding subspace $\mathcal V_r$, there exists:
\begin{equation}
d\big(\mathcal V_r(X),\mathcal V_r(X')\big)
\;\ge\;
d\big(\mathrm{Col}(G),\mathrm{Col}(S G)\big)
- C\,\frac{\|N\|_2+\|N'\|_2+\|R\|_2}{\delta_{\min}}.
\end{equation}
\end{enumerate}
\end{theorem}

\noindent
\textbf{Takeaways.} 
Phase tokenization is structurally invariant under the cycle pattern change $S$ and only subject to perturbations from noise and small day-to-day mismatches. 
In contrast, patch tokenization generally suffers from a non-vanishing structural offset. 
Hence, \emph{phase tokenization is more robust and consistent under cycle pattern drifts}.


\section{Methodology}
\label{sec:method}

Given the focus on periodicity, we adopt the channel-independent paradigm~\citep{nie2023patchtst, zeng2023dlinear} and omit the channel dimension throughout the remainder of this paper. The objective of forecasting is to predict the future trajectory $\mathbf{Y}\in\mathbb{R}^{L_{\text{out}}}$ from an input sequence $\mathbf{X}\in\mathbb{R}^{L_{\text{in}}}$, where $L_{\text{in}}$ and $L_{\text{out}}$ denote the input and output lengths, respectively. In the following sections, we describe the data preprocessing procedure, present the proposed network architecture, and finally analyze the computational complexity of the method.


\subsection{Data Pre-Processing}
\label{subsec:phase-embedding}
\fakeparagraph{Normalization and De-Normalization} 
Following \cite{kim2021reversible}, we normalize inputs with their estimated mean and standard deviation, and de-normalize predictions to the original scale.

\fakeparagraph{Phase Tokenization and De-Tokenization}  
Phase tokenization transforms the one-dimensional input sequence into a two-dimensional phase--period matrix for the following processing. 
Conversely, phase de-tokenization reconstructs the predicted phase--period matrix back into a one-dimensional output sequence. 
Let $L_{\text{phase}}$ denote the period length, which can be estimated using autocorrelation analysis. 

To ensure that the input sequence length is a multiple of $L_{\text{phase}}$, we circularly pad the sequence to length $P_{\text{in}}* L_{\text{phase}}$, where $P_{\text{in}} = \left\lceil \frac{L_{\text{in}}}{L_{\text{phase}}} \right\rceil.$ 
As illustrated in \figref{fig:overview}, the padded sequence $\mathbf{X}$ is then reshaped into a phase--period matrix $\mathbf{X}_{\text{phase}} \in \mathbb{R}^{L_{\text{phase}} \times P_{\text{in}}}$, where each entry $\mathbf{X}_{\text{phase}}[\ell, p]$ corresponds to the observation at the $\ell^\text{th}$ phase of the $p^\text{th}$ period. 
In the de-tokenization process, the predicted phase–period matrix is mapped back to the temporal domain by reversing the transformation, thereby reconstructing the final one-dimensional forecast sequence.

\subsection{Phase-based Routing Transformer}
\label{subsec:phase-network}

\begin{figure}
    \centering
    \includegraphics[width=0.95\linewidth]{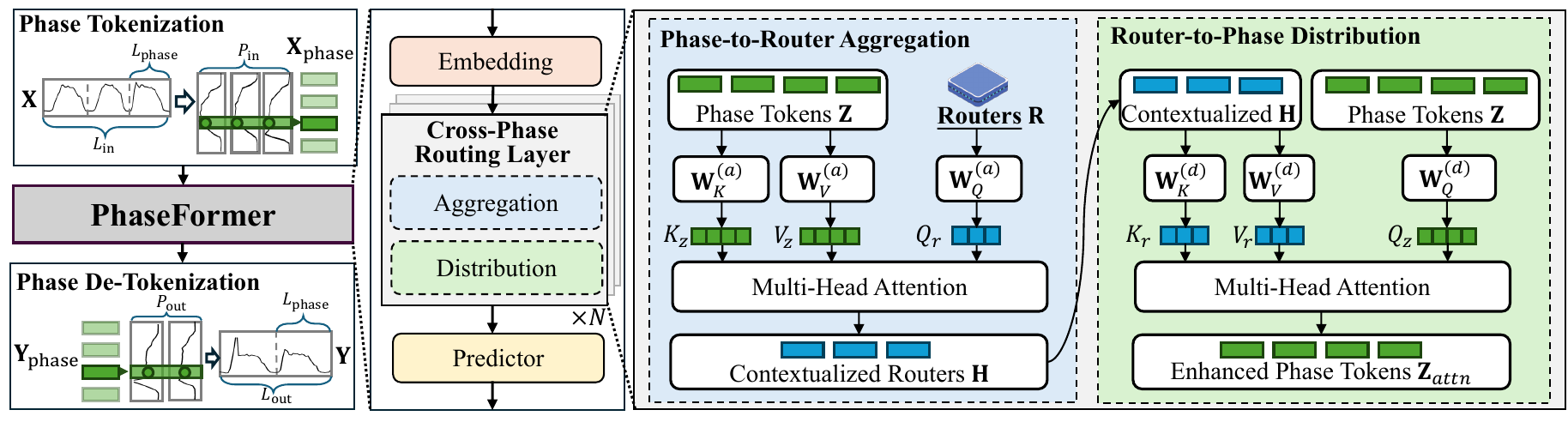}
    \caption{The overview of \sysname.}
    \label{fig:overview}
\end{figure}

The phase--period matrix is fed into our proposed phase-based routing Transformer, termed \sysname, to capture and extrapolate temporal dynamics at the phase level in an efficient and effective way. 
As illustrated in \figref{fig:overview}, \sysname first applies an embedding layer to the phase tokens, then refines them through multiple cross-phase routing layers, and finally maps them to the target via a shared predictor. Next, we elaborate on the design of these modules in detail.

\subsubsection{Embedding Layer}
\label{subsubsec:phase-embedding}
The embedding layer projects the phase tokens $\mathbf{X}_{\text{phase}}$ into a low-dimensional representation space, allowing the informative components to be extracted from raw observations that are often contaminated by perturbations. 
Formally, for each phase index $\ell \in \{1,\dots,L_{\text{phase}}\}$, the corresponding phase token $\mathbf{X}_{\text{phase}}[\ell, :]$ is mapped into a $d$-dimensional representation through a linear function $f_{\theta}$, parameterized by $\theta \in \mathbb{R}^{P_{\text{in}}\times d}$:
\begin{equation}
\mathbf{Z} 
= f_{\theta}\!\left(\mathbf{X}_{\text{phase}}\right)\in \mathbb{R}^{L_{\text{phase}}\times d}
\label{eq:phase-embedding}
\end{equation}

To better capture the temporal ordering among phases, we introduce a set of learnable positional embeddings $\mathbf{E}_{\text{pos}}\in\mathbb{R}^{L_{\text{phase}}\times d}$ to distinguish the relative position of each phase, following \cite{liu2023spatio}. 
These embeddings are added to $\mathbf{Z}$ in a phase-wise manner, so that each phase representation is enriched with its positional information:
\begin{equation}
\tilde{\mathbf{Z}}=\mathbf{Z}+\mathbf{E}_{\text{pos}}.
\end{equation}
The resulting $\tilde{\mathbf{Z}}$ is then forwarded to the cross-phase routing layers for higher-level feature interaction and forecasting.

\subsubsection{Cross-Phase Routing Layer}
Directly modeling full pairwise interactions among phase representations via self-attention is computationally expensive. To handle this, we introduce a set of learnable routers $\mathbf{R}\in\mathbb{R}^{M\times d}$ to mediate information exchange across phases, drawing inspiration from previous methods~\citep{jaegle2021perceiver, zhang2023crossformer}. This design substantially reduces the quadratic cost of self-attention while preserving rich cross-phase dependencies.

Cross-phase routing consists of two steps:
(i) \emph{phase-to-router aggregation}, which selectively compresses information from phase representations into the compact set of routers; and (ii) \emph{router-to-phase distribution}, which selectively propagates the aggregated cross-phase information from the routers back to the phase representations. Both steps are implemented via cross-attention, allowing the model to scale efficiently while preserving strong representational capacity.


\fakeparagraph{Phase-to-Router Aggregation}  
The routers attend to the phase representations to extract contextual information, yielding contextualized router embeddings 
$\mathbf{H}\in\mathbb{R}^{M\times d}$.  
Specifically, the routers act as queries while the phases provide keys and values. The projection matrices 
$\mathbf{W}_Q^{\text{agg}},\mathbf{W}_K^{\text{agg}},\mathbf{W}_V^{\text{agg}}\in\mathbb{R}^{d\times d}$ 
map the representations into query, key, and value spaces, respectively:  
\begin{equation}
\mathbf{Q}_r=\mathbf{R}\mathbf{W}_Q^{\text{agg}},\quad
\mathbf{K}_z=\tilde{\mathbf{Z}}\mathbf{W}_K^{\text{agg}},\quad
\mathbf{V}_z=\tilde{\mathbf{Z}}\mathbf{W}_V^{\text{agg}}.
\end{equation}
The aggregated router embeddings are then obtained via multi-head attention (MHA) with $d_h$ heads:  
\begin{equation}
\mathbf{H}=\mathrm{MHA}\!\left(\mathbf{Q}_r,\mathbf{K}_z,\mathbf{V}_z\right).
\end{equation}

\fakeparagraph{Router-to-Phase Distribution}  
The aggregated information in the routers is subsequently redistributed to the phase representations, thereby enabling cross-phase information flow. In this step, the phase representations serve as queries while the routers provide keys and values, yielding refined phase representations $\mathbf{Z}_{\text{attn}}$. The projection matrices $\mathbf{W}_Q^{\text{dist}}, \mathbf{W}_K^{\text{dist}}, \mathbf{W}_V^{\text{dist}}\in\mathbb{R}^{d\times d}$ are used for this distribution:  
\begin{equation}
\mathbf{Q}_z=\tilde{\mathbf{Z}}\mathbf{W}_Q^{\text{dist}},\quad
\mathbf{K}_r=\mathbf{H}\mathbf{W}_K^{\text{dist}},\quad
\mathbf{V}_r=\mathbf{H}\mathbf{W}_V^{\text{dist}},
\end{equation}
\begin{equation}
\mathbf{Z}_{\text{attn}}=\mathrm{MHA}\!\left(\mathbf{Q}_z,\mathbf{K}_r,\mathbf{V}_r\right).
\end{equation}
This mechanism restores phase-level resolution while simultaneously enforcing coherence across phases through the contextualized routers. Ultimately, each phase representation attends to all others through a two-stage routing pathway.

\subsubsection{Predictor}
\label{subsubsec:phase-predictor}

The predictor produces multi-step forecasts of length $P_{\text{out}}$ for all phases simultaneously, based on their refined representations. Taking as input the refined phase representations $\mathbf{Z}_{\text{attn}} \in \mathbb{R}^{L_{\text{phase}}\times d}$ from the final cross-phase routing layer, the predictor is realized as a linear mapping $g_{\phi}$, parameterized by $\phi \in \mathbb{R}^{d \times P_{\text{out}}}$:  
\begin{equation}
\mathbf{Y}_{\text{phase}} = g_{\phi}(\mathbf{Z}_{\text{attn}}) 
\in \mathbb{R}^{L_{\text{phase}}\times P_{\text{out}}}.
\label{eq:phase-predictor}
\end{equation}
All phases share the same predictor parameters, which enforces consistency across phases and reduces the number of trainable parameters. This not only improves efficiency but also regularizes learning, thereby enhancing generalization. Finally, the predicted phase--period matrix $\mathbf{Y}_{\text{phase}}$ is passed through de-tokenization and de-normalization to produce the final forecast $\mathbf{Y}$.

\subsection{Complexity of \sysname}

For each variable, the overall complexity of \sysname\ can be summarized as follows:  
the phase embedding layer requires $O(L_{\text{phase}} P_{\text{in}} d)$ time and $O(L_{\text{phase}} d)$ memory.  
The cross-phase routing layer, which dominates computation, incurs $O((L_{\text{phase}}+M)d^2 + M L_{\text{phase}} d)$ time and $O(H M L_{\text{phase}} + (L_{\text{phase}}+M)d)$ memory.  
Finally, the predictor costs $O(L_{\text{phase}} d P_{\text{out}})$ time and $O(L_{\text{phase}} P_{\text{out}})$ memory.  
Aggregating across \(N\) blocks, the end-to-end time complexity is:
\[
O\Big(N\big((L_{\text{phase}}+M)d^2 + M L_{\text{phase}} d\big)
+ L_{\text{phase}}d(P_{\text{in}}+P_{\text{out}})\Big).
\]
Substituting $P_{\text{in}}=\lceil L_{\text{in}}/L_{\text{phase}}\rceil$ and 
$P_{\text{out}}=\lceil L_{\text{out}}/L_{\text{phase}}\rceil$ into the above expression gives:
\[
O\Big(N\big((L_{\text{phase}}+M)d^2 + M L_{\text{phase}} d\big)
+ d(L_{\text{in}}+L_{\text{out}})\Big).
\]
As investigated in \secref{sec:motivations}, the phase token space exhibits a inherently low-dimensional structure, which allows $M$ and $d$ to be chosen as fixed and small numbers. 
Thus, the computational cost grows in a linear manner with both the input length $L_{\text{in}}$ and the output horizon $L_{\text{out}}$.
\section{Experiments}
\label{sec:exp}

\subsection{Long-term Time Series Forecasting}

We conduct a joint evaluation of model efficiency and predictive accuracy. 
The comparative analysis highlights that the proposed \textit{\sysname} establishes an improved effectiveness-efficiency tradeoff in terms of parameter scale and error metrics.
We also provide the code in \url{https://github.com/neumyor/PhaseFormer_TSL}.

\fakeparagraph{Datasets and Setup}
Experiments are performed on seven widely used long-term time series forecasting datasets: 
\textit{ETTh1}, 
\textit{ETTh2}, 
\textit{ETTm1}, 
\textit{ETTm2}\footnote{\url{https://github.com/zhouhaoyi/ETDataset}}, 
\textit{Weather}\footnote{\url{https://www.bgc-jena.mpg.de/wetter/}}, 
\textit{Electricity}\footnote{\url{https://archive.ics.uci.edu/ml/datasets}}, 
and \textit{Traffic}\footnote{\url{https://pems.dot.ca.gov/}}, 
covering a diverse range of real-world scenarios.
The details of the datasets are provided in \secref{sec:dataset_details}. 
Following prior works~\citep{nie2023patchtst, zhang2023crossformer, huang2025timebase}, we adopt a 6:2:2 split for the ETT datasets and a 7:1:2 split for the other datasets. 
For \sysname, we report the average results over three random seeds, while for the other baselines we follow their official implementations and released code. 
We evaluate the forecasting \textbf{accuracy} of all tested models using mean squared error (MSE) and mean absolute error (MAE), 
and assess \textbf{efficiency} in terms of floating-point operations (FLOPs) and the number of parameters (Params).

\fakeparagraph{Baselines and Implementation Details}
We evaluate our approach against eight competitive baselines, encompassing both state-of-the-art Transformer architectures and efficiency-oriented forecasting models.  
We compare our method with {PatchTST}\citeyearpar{nie2023patchtst}, {iTransformer}\citeyearpar{liu2023itransformer}, {Crossformer}\citeyearpar{zhang2023crossformer}, {FEDformer}\citeyearpar{zhou2022fedformer}, {TimeBase}\citeyearpar{huang2025timebase}, {SparseTSF}\citeyearpar{lin2024sparsetsf}, {FITS}\citeyearpar{fits2024}, and {TimeMixer}\citeyearpar{timemixer24}.
Among these, PatchTST, Crossformer, and TimeMixer are patch-based; SparseTSF is phase-based; TimeBase integrates patch and phase paradigms; FITS and FEDformer are frequency-domain; and iTransformer models the full sequence directly.
For all baselines, we adopt the recommended configurations provided in their official implementations.
The model is optimized using the Adam optimizer with a fixed learning rate of $1 \times 10^{-3}$.  
Following the settings from efficiency-oriented works~\citep{huang2025timebase, lin2024sparsetsf, fits2024}, the look-back length is set to 720 time steps.
More implementation details are provided in \secref{sec:dataset_details}.

\begin{table}[t]
\centering
\caption{
Main results for long-term forecasting.
The input sequence length is $L_{\text{input}} = 720$, and results are averaged over forecast horizons $L_{\text{out}} \in \{96, 192, 336, 720\}$.
The best results are shown in \best{bold}, and the second-best in \second{underline}.
}
\label{tab:main_results}
\resizebox{\textwidth}{!}{
\begin{tabular}{c|cc|cc|cc|cc|cc|cc|cc|cc|cc}
\hline
\multirow{2}{*}{Dataset}
& \multicolumn{2}{c|}{\sysname} 
& \multicolumn{2}{c|}{PatchTST} 
& \multicolumn{2}{c|}{iTransformer} 
& \multicolumn{2}{c|}{Crossformer}
& \multicolumn{2}{c|}{FEDformer}
& \multicolumn{2}{c|}{TimeBase} 
& \multicolumn{2}{c|}{SparseTSF} 
& \multicolumn{2}{c|}{FITS} 
& \multicolumn{2}{c}{TimeMixer} \\ 
\cline{2-19}
 & MSE & MAE & MSE & MAE & MSE & MAE & MSE & MAE & MSE & MAE & MSE & MAE & MSE & MAE & MSE & MAE & MSE & MAE \\
\hline
ETTh1 & \best{0.403} & \best{0.415} & 0.420 & 0.439 & 0.453 & 0.467 & 0.517 & 0.512 & 0.523 & 0.523 & \second{0.404} & \second{0.416} & 0.406 & 0.418 & 0.419 & 0.435 & 0.452 & 0.474 \\
ETTh2 & 0.346 & 0.388 & 0.344 & 0.390 & 0.392 & 0.422 & 1.468 & 0.867 & 0.428 & 0.469 & 0.347 & 0.397 & \second{0.345} & \second{0.383} & \best{0.334} & \best{0.382} & 0.386 & 0.425 \\
ETTm1 & \best{0.346} & \best{0.374} & \second{0.354} & 0.383 & 0.370 & 0.401 & 0.390 & 0.417 & 0.438 & 0.465 & 0.356 & \second{0.380} & 0.362 & 0.383 & 0.359 & 0.382 & 0.383 & 0.413 \\
ETTm2 & \best{0.250} & \best{0.313} & 0.251 & 0.319 & 0.278 & 0.337 & 0.392 & 0.426 & 0.401 & 0.452 & \best{0.250} & \second{0.314} & 0.252 & 0.316 & 0.285 & 0.336 & 0.314 & 0.367 \\
Electricity & \best{0.160} & \best{0.250} & 0.169 & 0.265 & \second{0.165} & 0.263 & 0.180 & 0.273 & 0.235 & 0.348 & 0.167 & \second{0.258} & 0.168 & 0.263 & 0.172 & 0.270 & 0.171 & 0.273 \\
Traffic & \best{0.386} & \best{0.249} & \second{0.394} & \second{0.266} & 0.406 & 0.290 & 0.545 & 0.282 & 0.638 & 0.400 & 0.418 & 0.278 & 0.413 & 0.280 & 0.410 & 0.290 & 0.421 & 0.298 \\
Weather & \best{0.223} & \best{0.260} & \second{0.223} & 0.264 & 0.233 & 0.273 & 0.255 & 0.304 & 0.354 & 0.393 & 0.227 & \second{0.262} & 0.243 & 0.285 & 0.241 & 0.283 & 0.237 & 0.281 \\
\hline
\end{tabular}}
\end{table}

\begin{figure}[t]
    \centering
    \begin{subfigure}{0.24\textwidth}
        \centering
        \includegraphics[width=\linewidth]{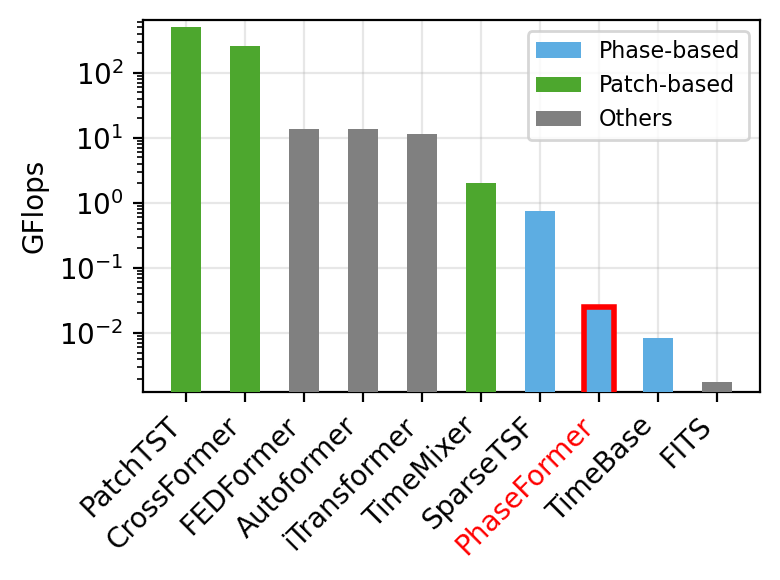}
        \caption{FLOPs on Traffic}
        \label{fig:traffic_flops}
    \end{subfigure}
    \begin{subfigure}{0.24\textwidth}
        \centering
        \includegraphics[width=\linewidth]{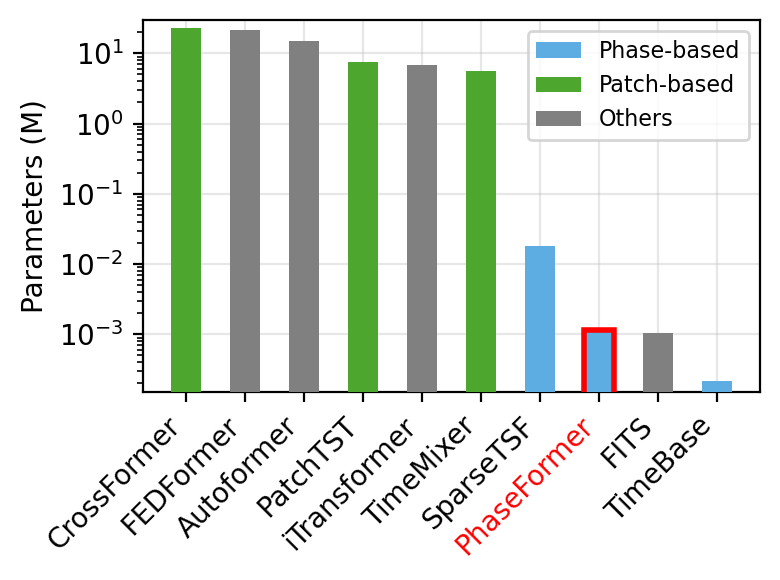}
        \caption{Params on Traffic}
        \label{fig:traffic_params}
    \end{subfigure}
    \begin{subfigure}{0.24\textwidth}
        \centering
        \includegraphics[width=\linewidth]{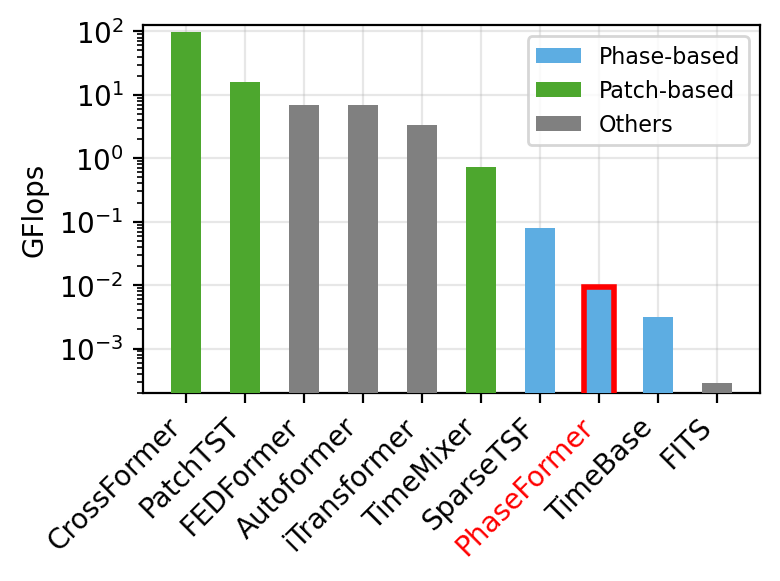}
        \caption{FLOPs on Electricity}
        \label{fig:electricity_flops}
    \end{subfigure}
    \begin{subfigure}{0.24\textwidth}
        \centering
        \includegraphics[width=\linewidth]{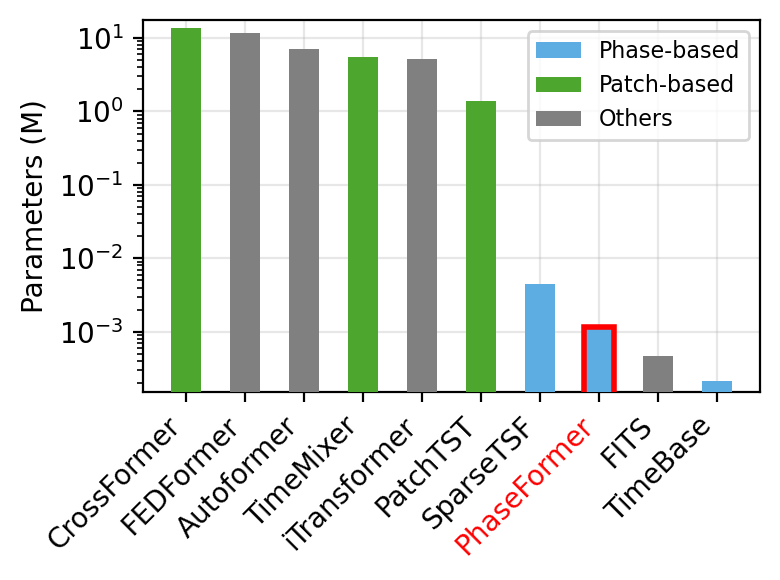}
        \caption{Params on Electricity}
        \label{fig:electricity_params}
    \end{subfigure}
    \caption{
    Comparison of FLOPs and parameter counts across models on the Traffic and Electricity.
    Patch-based models are shown in green, phase-based models in blue, and other models in gray.
    }
    \label{fig:flops_params_visualization}
\end{figure}

\fakeparagraph{Main Results}
We evaluate the predictive accuracy of \sysname and the baseline methods on seven datasets. 
\tabref{tab:main_results} reports the average prediction errors across four forecasting horizons, with detailed results provided in \secref{sec:accuracy_details}. 
Overall, \sysname consistently achieves superior performance on nearly all datasets, with particularly notable gains on complex and dynamic benchmarks such as Weather, Electricity, and Traffic.
For example, on the largest dataset, Traffic, \sysname surpasses the second-best method, PatchTST, by 6.3\% and outperforms TimeBase by 10.4\%, underscoring its robustness on large-scale and heterogeneous data. 
The only exception is ETTh2, where \sysname ranks second to FITS while still maintaining highly competitive accuracy.
A closer examination reveals that patch-based baselines, including PatchTST, Crossformer, and TimeMixer, exhibit performance degradation on the Electricity, likely due to stronger distributional shifts. 
In contrast, \sysname delivers stable and consistently superior results under these challenging conditions, highlighting the robustness of its phase-based design.

\fakeparagraph{Efficiency Comparison}
We evaluate the computational overhead of all models, with detailed results in \secref{sec:efficiency_details}.
\figref{fig:flops_params_visualization} shows the FLOPs and the number of parameters of all tested models on the Electricity and Traffic.
Overall, phase-based models incur lower overhead than patch-based ones.
On the Traffic dataset, \sysname achieves an extraordinary FLOPs reduction of about 99.99\% over PatchTST and Crossformer.
Beyond patch-based baselines, it also outperforms other phase-based models like SparseTSF, consistently delivering high efficiency.
This stems from the lower variety of phase tokens over time (\secref{sec:motivations}), making them inherently more efficient to process. 
Taken together with the previous accuracy evaluations, these results clearly demonstrate that \sysname provides an efficient yet effective solution, delivering superior performance on complex datasets.

\subsection{Ablation Studies and Analysis}

\begin{figure}[t]
    \centering
    \begin{subfigure}{0.32\textwidth}
        \centering
        \includegraphics[width=\linewidth]{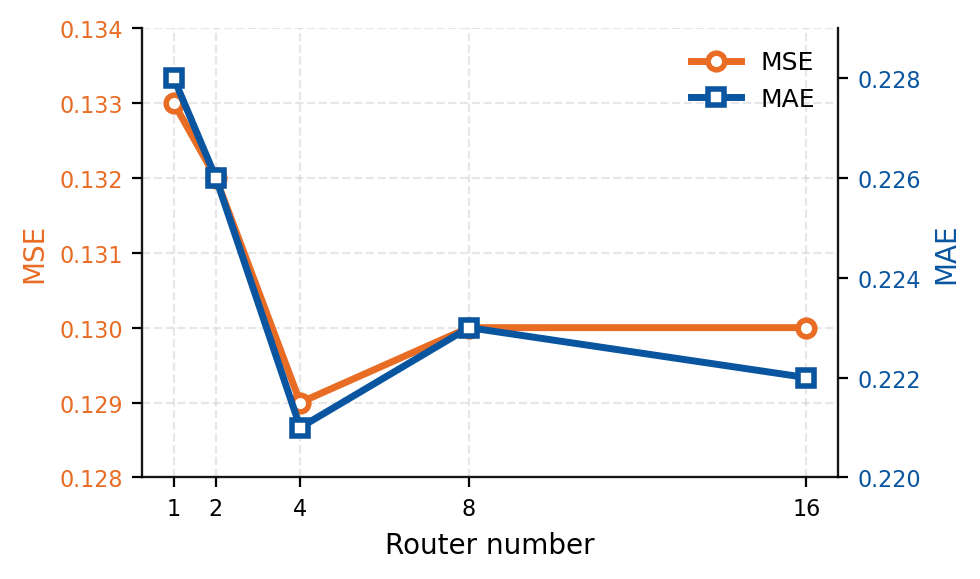}
        \caption{Electricity}
        \label{fig:electricity}
    \end{subfigure}
    \begin{subfigure}{0.32\textwidth}
        \centering
        \includegraphics[width=\linewidth]{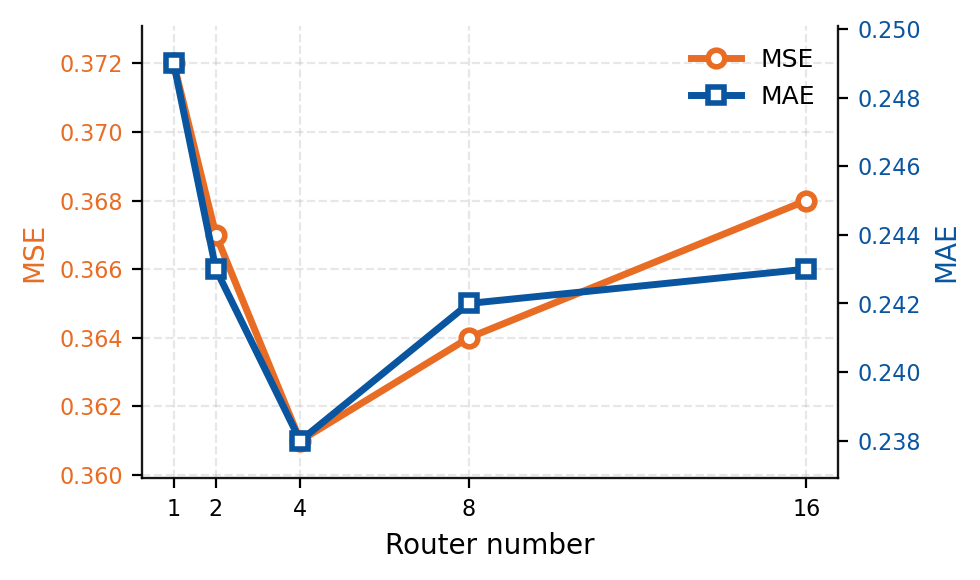}
        \caption{Traffic}
        \label{fig:traffic}
    \end{subfigure}
    \begin{subfigure}{0.32\textwidth}
        \centering
        \includegraphics[width=\linewidth]{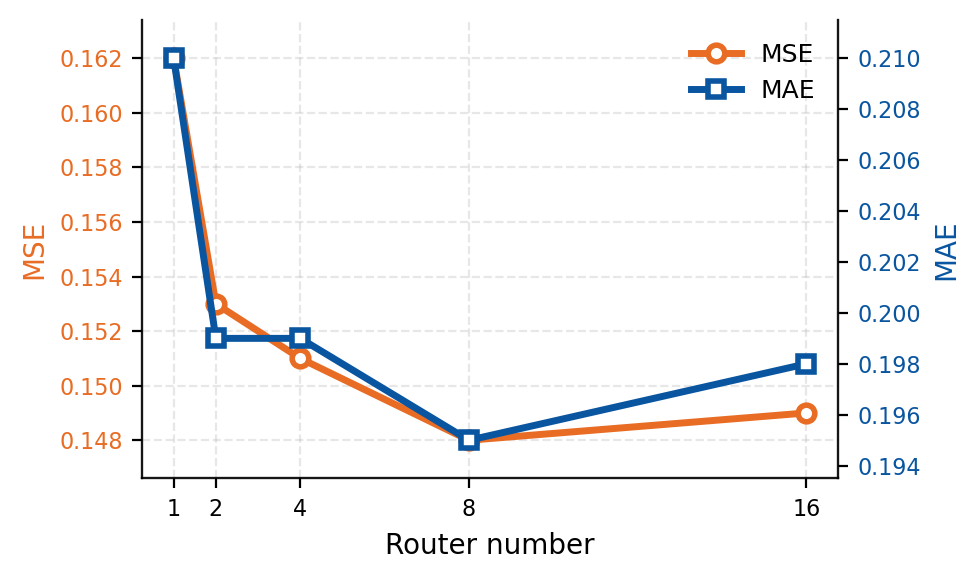}
        \caption{Weather}
        \label{fig:Weather}
    \end{subfigure}
    
    \caption{Effect of varying the number of routers $M$ on forecasting performance on three datasets.}
    \label{fig:router_number_visualization}
\end{figure}

\fakeparagraph{Varying the Number of Routers}
We systematically evaluate the impact of different number of routers $M$ on model performance, with results summarized in \figref{fig:router_number_visualization}.
The experiments indicate that across three datasets, the model’s prediction error generally decreases as the number of routers $M$ increases, before eventually stabilizing or slightly rising.
It is worth noting that the best performance is usually achieved when $M \in \{4,8\}$, which is much smaller compared to the actual number of phase tokens, $L_{\text{phase}} = 24$.
This observation indicates that the phase token spans an inherently low-dimensional space, so only a small number of routers is sufficient to effectively capture and represent its underlying structure.
More detailed results are provided in \secref{sec:all_results_router_numbers}.

\begin{table}[t]
\centering
\caption{Cross-Phase Routing layer ablation. Each cell reports MSE, MAE, and FLOPs. Lower is better for all metrics. FLOPs are reported in millions (MFLOPs). The best results are highlighted with \best{Bold}, and the second-best results with \second{Underlined}.}
\label{tab:ablation_phase}
\resizebox{\textwidth}{!}{
\begin{tabular}{c|ccc|ccc|ccc|ccc}
\hline
\multirow{2}{*}{Dataset}
& \multicolumn{3}{c|}{\textbf{\sysname}} 
& \multicolumn{3}{c|}{\textbf{w/ FullAttention}} 
& \multicolumn{3}{c|}{\textbf{w/ LinearMixing}} 
& \multicolumn{3}{c}{\textbf{w/o Routing}} \\
\cline{2-13}
& MSE & MAE & FLOPs 
& MSE & MAE & FLOPs 
& MSE & MAE & FLOPs 
& MSE & MAE & FLOPs \\
\hline
Weather 
& \best{0.1503} & \best{0.1971} & 3.119 
& \second{0.1527} & \second{0.2005} & 3.202 
& 0.1700 & 0.2226 & \second{0.920} 
& 0.1907 & 0.2406 & \best{0.783} \\
\hline
Electricity 
& \best{0.1290} & \best{0.2209} & 42.213 
& \second{0.1295} & \second{0.2217} & 48.951 
& 0.1403 & 0.2334 & \second{14.068} 
& 0.1423 & 0.2365 & \best{11.972 } \\
\hline
Traffic     
& \best{0.3721} & \best{0.2475} & 113.356 
& \second{0.3791} & \second{0.2513} & 131.452 
& 0.3842 & 0.2532 & \second{37.776}
& 0.3892 & 0.2584 & \best{32.149}  \\
\hline
\end{tabular}}
\end{table}

\fakeparagraph{Effectiveness of Cross-Phase Routing} 
To assess the contribution of the cross-phase routing layer, we compare four variants of the model: 
\textbf{\sysname}, which adopts the original cross-phase routing layer; 
\textbf{w/ FullAttention}, which substitutes the cross-phase routing layer with a full attention mechanism; and
\textbf{w/ LinearMixing}, which replaces the cross-phase routing layer with a linear layer; and 
\textbf{w/o Routing}, which directly projects each phase into its own future.
All other experimental settings are kept identical across these variants. 

As summarized in \tabref{tab:ablation_phase}, \sysname consistently outperforms \textbf{w/ LinearMixing} and \textbf{w/o Routing}, indicating that explicit cross-phase routing is crucial for modeling periodic dynamics. Moreover, \sysname not only incurs less computational and memory overhead, but also achieves lower prediction error than \textbf{w/ FullAttention}, showing that the routing layer is both efficient and effective. We attribute these gains to operating in a low-dimensional phase token space, which concentrates informative interactions and reduces cost.

\begin{figure}[t] 
  \centering
  \begin{subfigure}{0.24\linewidth}
    \includegraphics[width=\linewidth]{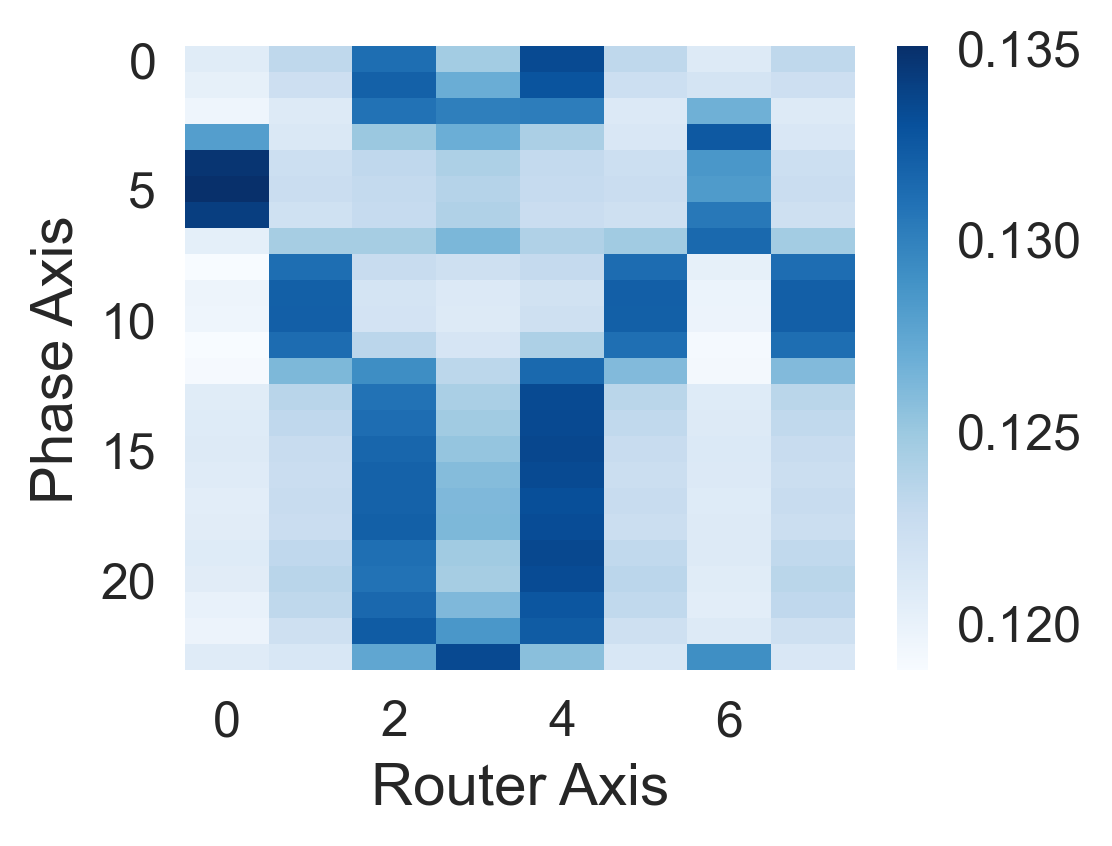}
    \caption{Aggregation Weights}
    \label{fig:sender}
  \end{subfigure}%
  \hfill
  \begin{subfigure}{0.24\linewidth}
    \includegraphics[width=\linewidth]{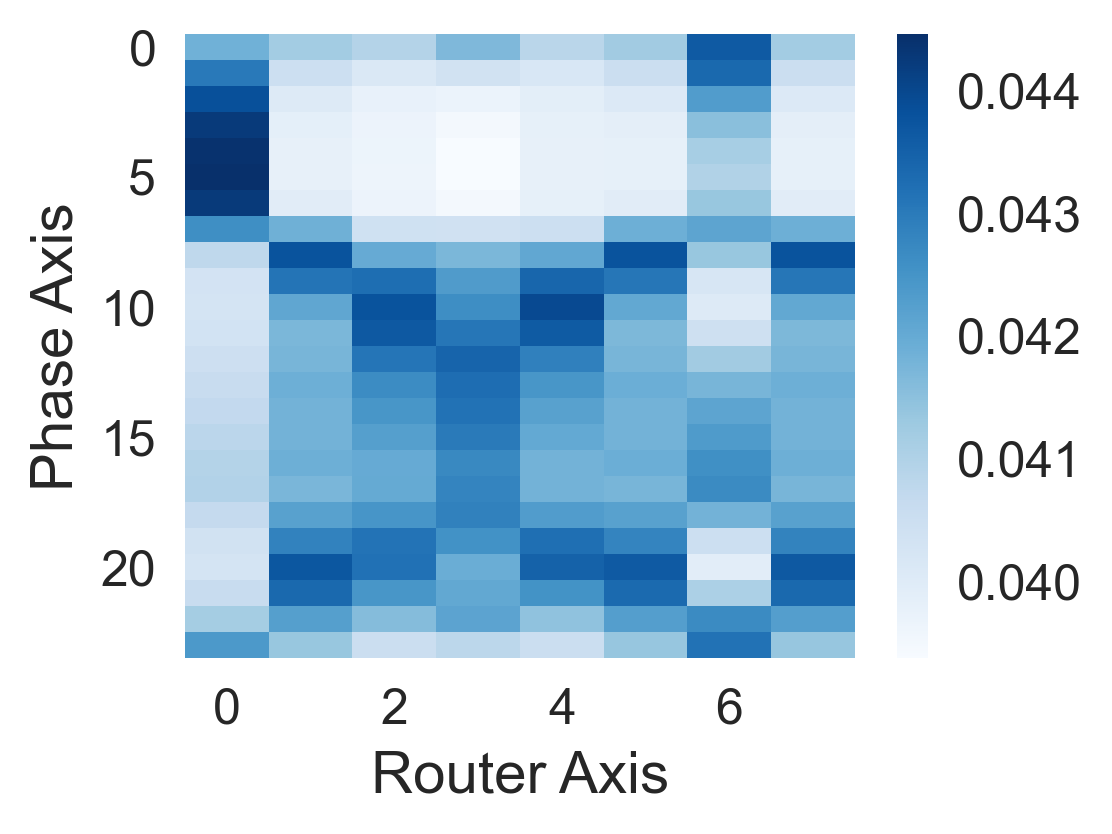}
    \caption{Distribution Weights}
    \label{fig:receiver}
  \end{subfigure}
  \hfill
  \begin{subfigure}{0.48\linewidth}
    \includegraphics[width=\linewidth]{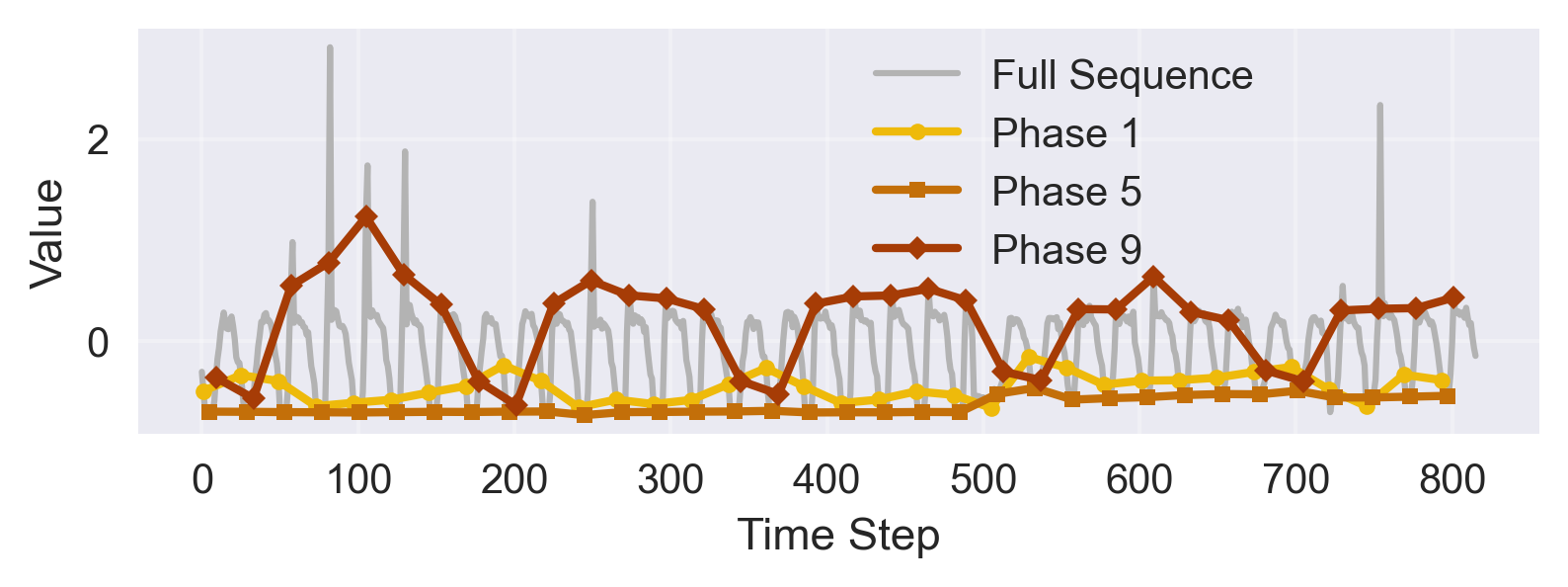}
    \caption{Visualization of three selected phase tokens}
    \label{fig:traffic_data_visualization}
  \end{subfigure}%
  \caption{Case study on a sample from the Traffic dataset. 
  (a) Attention weight matrix during Phase-to-Router aggregation. 
  (b) Attention weight matrix during Router-to-Phase distribution. 
  Both matrices capture the association between 8 routers and 24 input phases. 
  (c) Visualization of three representative phases (1, 5, and 9), each represents a distinct attentive pattern with routers.} 
  \label{fig:case_study_wrap_all}
\end{figure}

\subsection{Case Study}
We select one sample from Traffic dataset, comprising an input sequence of length 720 and an output sequence of length 96 (816 time steps in total). 
The input sequence is fed into \sysname, and we record the attention-weight matrices at the first cross-phase routing layer during both aggregation (Phase$\rightarrow$Router) and distribution (Router$\rightarrow$Phase).
As shown in \figref{fig:sender} and \figref{fig:receiver}, both attention patterns exhibit clear local similarity: adjacent phases tend to be assigned to the same routers and to receive attention from similar routers.
This indicates that the routing mechanism captures temporally consistent phase relationships. 
Meanwhile, the attention weights reveal that certain phases share similar attentive patterns.
To analyze this further, we focus on three phases with distinct attentive patterns and visualized them in \figref{fig:traffic_data_visualization}.
These phases display different temporal behaviors: Phase 5 remains relatively stable over long horizons, whereas Phase 9 and Phase 1 both exhibit a pronounced 7-day periodicity but with opposite trends.
The differing patterns of these phase tokens suggest that the router structure not only distinguishes among phase tokens but also effectively models their periodicity and trend characteristics.

\section{Conclusion}
\label{sec:conclusion}

This work identifies the inefficiencies of patch-based forecasting and presents \sysname, a phase-centric model that captures periodicity via compact phase representations and lightweight cross-phase routing. 
Both theoretical analysis and empirical validation converge on the same conclusion that phase representations remain both more robust and more efficient than patch-based approaches under cycle pattern shifts. 
Consequently, \sysname maintains high predictive accuracy while remaining lightweight compared to patch-based methods.
More broadly, these results provide a practical pathway for building lightweight yet powerful forecasting models that retain accuracy without heavy and complex architectures.

However, the approach assumes locally stable periodicity across the input and output horizons; under highly irregular or non-repetitive cycles, phase representations may fail to capture meaningful dynamics. Future work will relax this assumption by modeling non-stationarity and complex drifts, aiming to develop more resilient phase representations and further establish \sysname as a benchmark for long-term time-series forecasting.
\newpage
\section{Ethics Statement}

This study focuses on methodological advances in time-series forecasting and does not involve human subjects, personally identifiable data, or sensitive private information.
All experiments use publicly available benchmark datasets that are widely adopted in the research community, and their use complies with the terms of release.
We do not employ proprietary or confidential data, and no conflicts of interest exist.
The contributions are purely technical in nature and do not promote harmful applications.
All authors affirm adherence to fairness, research integrity, and relevant legal and ethical standards, in line with the ICLR Code of Ethics.

\section{Reproducibility Statement}

We make substantial efforts to ensure reproducibility. 
All datasets used in our experiments are publicly accessible, with links provided in \secref{sec:exp}. 
Detailed dataset statistics, preprocessing steps, and partitioning procedures appear in \secref{sec:dataset_details}. 
Model architectures, hyperparameters, and training procedures (including optimizer choice, learning rate, look-back window length, and router configuration) are described in \secref{sec:exp} and the \secref{sec:dataset_details}.

For fair comparison, we follow the official implementations of all baseline models and provide references to their sources. Comprehensive experimental results, including ablation studies, efficiency analyses, and visualizations, appear in \secref{sec:exp} and \secref{sec:full_results}. Theoretical analyses supporting our design choices also appear in the \secref{sub:theoretical}.

Finally, to facilitate independent verification, we release anonymized source code and experiment scripts in a public repository at \url{https://anonymous.4open.science/r/ICLR26-PhaseFormer-17678}. 
Collectively, these measures ensure that our reported results are reliably reproducible and extensible by the research community.
\bibliography{iclr2026_conference}
\bibliographystyle{iclr2026_conference}

\appendix
\section{Appendix}
\label{sec:appendix}

\subsection{Details about Baselines}
In our experiments, we incorporated a diverse set of time series forecasting models, with particular emphasis on approaches based on \textbf{Patch Tokenization} and efficient forecasting models. The details of these models are as follows:

\begin{enumerate}

    \item \textbf{PatchTST} — A channel-independent Transformer that treats each variable as an individual channel and segments the time series into patches as tokens. This design reduces the complexity of the attention mechanism and enables the utilization of longer historical sequences, thereby improving long-term forecasting accuracy.

    \item \textbf{iTransformer} — A channel-dependent Transformer that models variables themselves as tokens to capture inter-variable relationships, while simultaneously accounting for nonlinear temporal variations within each variable.

    \item \textbf{Crossformer} — A multi-scale Transformer that performs patching or segmentation along the temporal dimension and employs a two-stage attention mechanism (within-time and cross-variable). This design effectively captures both temporal dependencies and inter-variable correlations, making it particularly suitable for datasets characterized by strong inter-variable coupling and mixed long- and short-term patterns.

    \item \textbf{FEDformer} — A model that integrates trend-seasonal decomposition with frequency-domain analysis. It extracts a small number of significant frequency components to enhance periodic forecasting performance while maintaining controlled model complexity in long-term forecasting tasks.

    \item \textbf{SparseTSF} — A lightweight model that reduces temporal complexity through periodic down-sampling or subsequence selection, aiming to achieve competitive periodic forecasting performance with minimal resource consumption.

    \item \textbf{FITS} — A lightweight model that leverages frequency-domain features and interpolation operations to reconstruct the predicted sequences. With fewer parameters and low computational overhead compared with other models, it demonstrates strong performance on time series with distinct spectral structures.

    \item \textbf{TimeBase} — A model that constructs temporal bases (via patching or segmentation strategies) to represent historical and future variations. Its objective is to maintain satisfactory forecasting accuracy while reducing computational and parameter costs.

    \item \textbf{TimeMixer} — An patch-based forecasting model fully based on MLPs. It employs Past-Decomposable-Mixing to decouple seasonal and trend components across different scales (fine and coarse) and utilizes Future-Multipredictor-Mixing to aggregate multi-scale predictions. This design achieves a balance of efficiency and accuracy in both short-term and long-term forecasting tasks.

\end{enumerate}

\subsection{Implementation Details}
\label{sec:dataset_details}

\begin{table}[ht]
\centering
\begin{tabular}{lcccccc}
\hline
Dataset     & Var & Length & $T$ & $L$       & Freq   & Scale  \\
\hline
ETTh1       & 7   & 14,400 & 720 & 96$\sim$720 & 1hour   & 0.1M   \\
ETTh2       & 7   & 14,400 & 720 & 96$\sim$720 & 1hour   & 0.1M   \\
ETTm1       & 7   & 57,600 & 720 & 96$\sim$720 & 15mins  & 0.4M   \\
ETTm2       & 7   & 57,600 & 720 & 96$\sim$720 & 15mins  & 0.4M   \\
Weather     & 21  & 52,696 & 720 & 96$\sim$720 & 10mins  & 1.1M   \\
Electricity & 321 & 26,304 & 720 & 96$\sim$720 & 1hour   & 8.1M   \\
Traffic     & 862 & 17,544 & 720 & 96$\sim$720 & 1hour   & 15.0M  \\
\hline
\end{tabular}
\caption{Dataset statistics used in experiments.}
\label{tab:dataset_details}
\end{table}

We present detailed statistics of the datasets in \tabref{tab:dataset_details}. 
The data loading and preprocessing procedures follow prior works~\citep{nie2023patchtst, huang2025timebase}.

All baseline methods are implemented based on their original papers or official code. 
For cases where fixed random seeds are not specified, each experiment is repeated three times to ensure stability. All experiments are conducted using PyTorch~\citep{paszke2019pytorch} on a single NVIDIA A100 24GB GPU.

For model configuration, the primary period is determined via frequency-domain analysis by selecting the dominant component, while the number of routers is chosen through grid search.
We mainly use a single-layer model with 8 routers for ETT datasets, a two-layer model with 4 routers for Traffic and Electricity datasets, and a 3-layer model with 8 routers for the Weather dataset.
Please refer to the released code for complete training details at \url{https://github.com/neumyor/PhaseFormer_TSL}. 

\subsection{Full Results}
\label{sec:full_results}

\subsubsection{The Detailed Forecasting Accuracy Results}
\label{sec:accuracy_details}
We present detailed forecasting results across all prediction horizons on the test sets in \tabref{tab:full_results}, with the input length fixed to 720.
\sysname consistently delivers strong and stable performance across most datasets and forecasting lengths. The only notable exception is ETTh2, where FITS slightly outperforms our model. This highlights the robustness of \sysname across diverse scenarios, even though some simple datasets may still favor specialized baselines.
It is also worth noting that TimeBase, which adopts a phase-based strategy, achieves competitive results on the relatively simple ETT datasets. In contrast, \sysname demonstrates its advantage primarily on Traffic and Electricity, which are more complex and challenging datasets. This distinction illustrates that while phase-inspired models may be effective in straightforward settings, \sysname generalizes better and excels in more demanding real-world contexts.

\subsubsection{The Detailed Forecasting Efficiency Results}
\label{sec:efficiency_details}

We further provide the efficiency comparison of \sysname against all baselines in terms of FLOPs and number of parameters, with the input length set to 720 and the output length fixed at 96.
The results in \figref{tab:params_flops_details} reveal that \sysname achieves a favorable trade-off between accuracy and efficiency. Despite its stronger predictive performance, \sysname maintains moderate model size and computational cost, often comparable to or even lower than other transformer-based models:
On complex datasets such as Traffic, \sysname outperforms large baselines like PatchTST with substantially fewer FLOPs;
On simpler datasets, even when specialized models such as TimeBase or FITS show competitive accuracy, their efficiency advantage diminishes when considering scalability to larger, real-world datasets.
These findings underscore that \sysname is not only accurate but also efficient, making it more suitable for deployment in resource-constrained or latency-sensitive environments.

\subsubsection{The Detailed Results of PCA visualization}

\begin{figure}[t]
  \centering
  \begin{subfigure}[t]{0.32\textwidth}
    \centering
    \includegraphics[width=\linewidth]{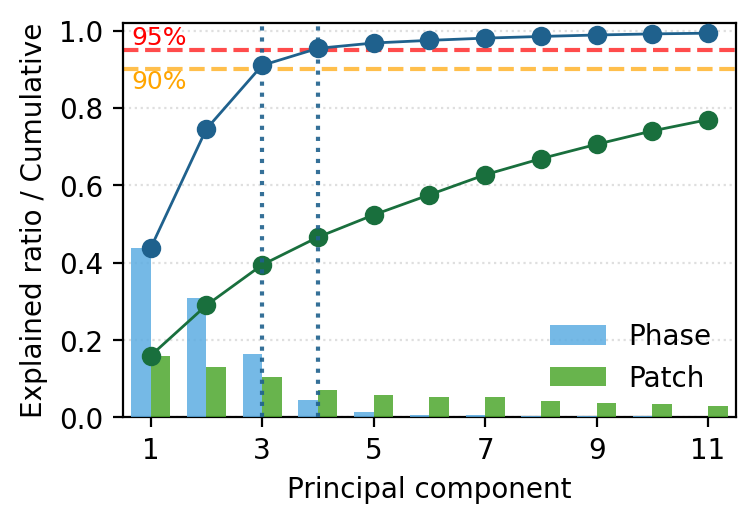}
    \caption{PCA for tokens on \emph{ETTh1}.}
    \label{fig:pca_etth1}
  \end{subfigure}
  \hfill
  \begin{subfigure}[t]{0.32\textwidth}
    \centering
    \includegraphics[width=\linewidth]{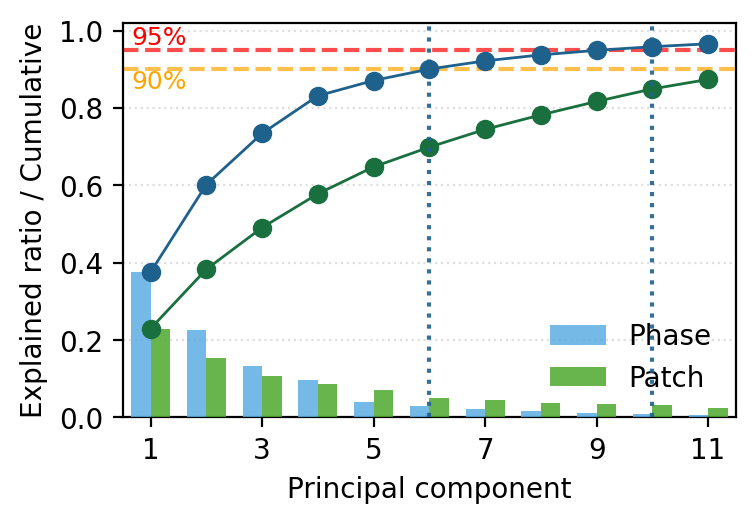}
    \caption{PCA for tokens on \emph{ETTh2}.}
    \label{fig:pca_etth2}
  \end{subfigure}
  \hfill
  \begin{subfigure}[t]{0.32\textwidth}
    \centering
    \includegraphics[width=\linewidth]{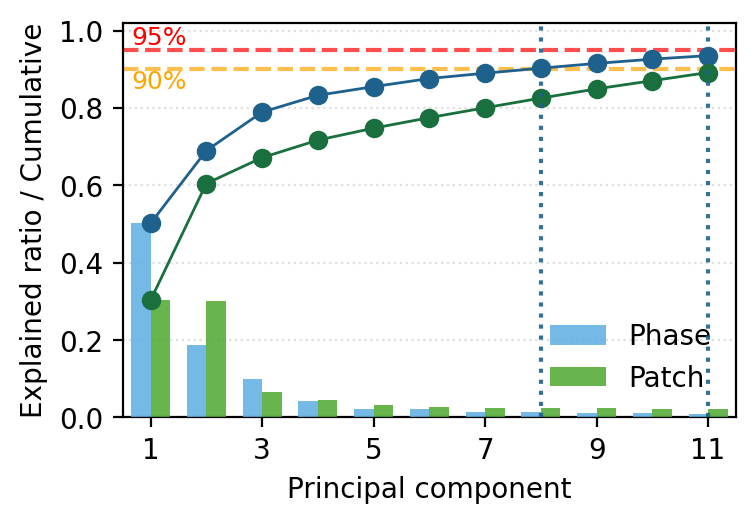}
    \caption{PCA for tokens on \emph{ETTm1}.}
    \label{fig:pca_ettm1}
  \end{subfigure}

  \vskip\baselineskip
  \begin{subfigure}[t]{0.32\textwidth}
    \centering
    \includegraphics[width=\linewidth]{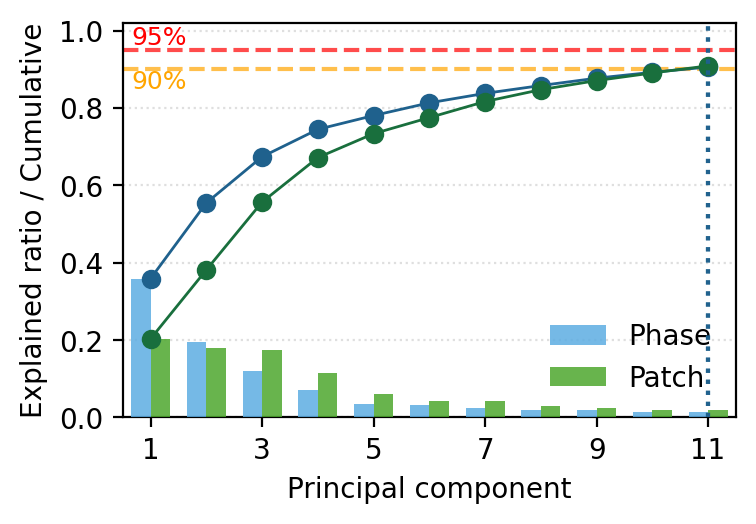}
    \caption{PCA for tokens on \emph{ETTm2}.}
    \label{fig:pca_ettm2}
  \end{subfigure}
  \hfill
  \begin{subfigure}[t]{0.32\textwidth}
    \centering
    \includegraphics[width=\linewidth]{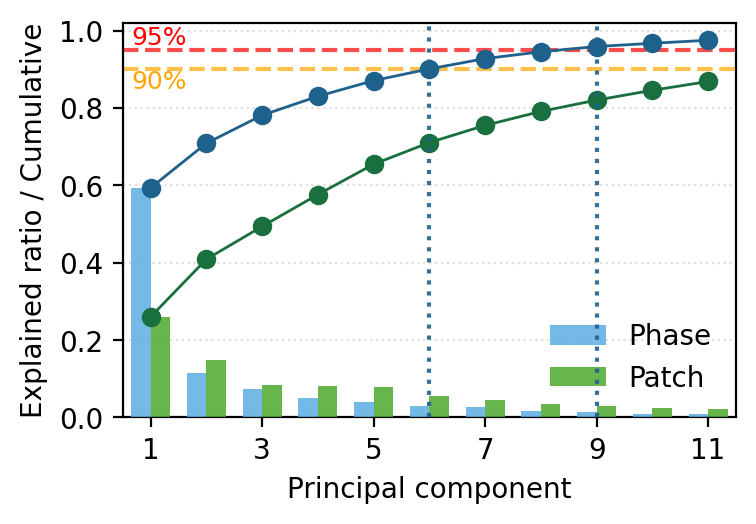}
    \caption{PCA for tokens on \emph{Electricity}.}
    \label{fig:pca_ecl}
  \end{subfigure}
  \hfill
  \begin{subfigure}[t]{0.32\textwidth}
    \centering
    \includegraphics[width=\linewidth]{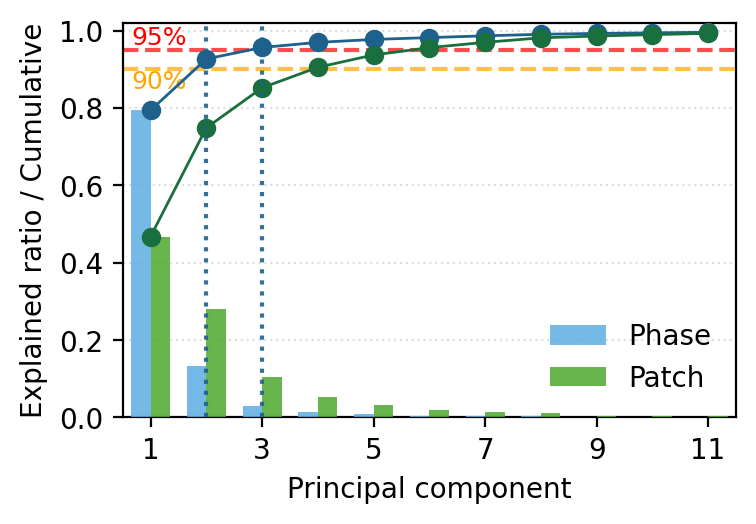}
    \caption{PCA for tokens on \emph{Weather}.}
    \label{fig:pca_weather}
  \end{subfigure}

  \caption{Visualization of phase tokenization across six datasets: ETTh1, ETTh2, ETTm1, ETTm2, Electricity, and Weather.}
  \label{fig:phase_token_comparison_detailed}
\end{figure}

We present PCA visualization results on the ETTh1, ETTh2, ETTm1, ETTm2, Electricity, and Weather datasets in Figure \ref{fig:phase_token_comparison_detailed}, in addition to Figure \ref{fig:traffic_pca_cycle_phase}.
The findings are consistent with those observed on the Traffic dataset: phase tokenization yields a significantly more compact space compared to patch tokenization.

\subsubsection{The Detailed Results of Varying Router Numbers}
\label{sec:all_results_router_numbers}
We further provide detailed results on the effect of varying the number of routers (1,2,4,8,16) across three datasets: Traffic, Electricity, and Weather. The input window was fixed at 720, and the output length was set to 96.

Our observations show that the number of routers does influence model performance, but the optimal configuration typically involves a relatively small number of routers. Specifically, the best performance was achieved with 8 routers on the Weather dataset, and with 4 routers on both the Electricity and Traffic datasets.
Since routers serve as the foundation for aggregation and distribution in the phase token space, these results provide supporting evidence that the phase token space captures low-dimensional features, allowing strong performance even with fewer routers.

\begin{table}[t]
\centering
\caption{Parameters and FLOPS across models for different datasets.}
\label{tab:params_flops_details}
\resizebox{\textwidth}{!}{
\begin{tabular}{c|cc|cc|cc|cc|cc|cc|cc}
\hline
\multirow{2}{*}{Model}  & \multicolumn{2}{c|}{Traffic} & \multicolumn{2}{c|}{Weather} & \multicolumn{2}{c|}{Electricity} & \multicolumn{2}{c|}{ETTh1} & \multicolumn{2}{c|}{ETTh2} & \multicolumn{2}{c|}{ETTm1} & \multicolumn{2}{c}{ETTm2} \\
\cline{2-15}
 & Params & FLOPS & Params & FLOPS & Params & FLOPS & Params & FLOPS & Params & FLOPS & Params & FLOPS & Params & FLOPS \\
\hline
\sysname   & 1.156K & 13.9 & 308 & 0.15 & 1.156K & 5.18 & 1.156K & 0.11 & 1.156K & 0.11 & 1.156K & 0.11 & 1.156K & 0.11 \\
PatchTST      & 7.589M & 498{,}577.49 & 1.373M & 1{,}054.77 & 1.373M & 16{,}122.93 & 587.68K & 51.29 & 587.68K & 51.29 & 587.68K & 51.29 & 587.68K & 51.29 \\
iTransformer  & 6.731M & 11{,}652.34 & 5.153M & 257.54 & 5.153M & 3{,}347.97 & 369.9K & 8.12 & 304.1K & 6.68 & 304.1K & 7.29 & 304.1K & 7.29 \\
Crossformer   & 22.954M & 259{,}209.90 & 158.34K & 84.09 & 13.537M & 96{,}564.63 & 2.069M & 544.20 & 2.069M & 544.20 & 2.069M & 544.20 & 2.069M & 544.20 \\
FEDformer     & 21.206M & 13{,}679.70 & 5.828M & 2{,}757.24 & 11.861M & 6{,}904.61 & 5.792M & 2{,}734.51 & 5.792M & 2{,}734.51 & 5.793M & 2{,}734.95 & 5.793M & 2{,}734.95 \\
TimeBase      & 214 & 8.44 & 214 & 0.21 & 214 & 3.14 & 214 & 0.07 & 214 & 0.07 & 704 & 0.23 & 704 & 0.23 \\
SparseTSF     & 17.949K & 751.31 & 4.509K & 5.14 & 4.509K & 78.61 & 4.509K & 1.71 & 4.509K & 1.71 & 4.509K & 1.71 & 4.509K & 1.71 \\
FITS          & 1.054K & 1.76 & 272 & 0.01 & 462 & 0.28 & 272 & 0.004 & 272 & 0.004 & 2.646K & 0.04 & 2.646K & 0.04 \\
TimeMixer     & 5.697M & 2{,}026.53 & 5.562M & 205.40 & 5.584M & 739.64 & 4.024M & 125.91 & 4.024M & 125.91 & 4.024M & 125.95 & 4.024M & 125.95 \\
\hline
\end{tabular}}
\end{table}

\begin{table}[t]
\centering
\caption{Full results across datasets and prediction lengths. 
Each entry reports MAE and MSE. The input length is set to 720. The best results are marked with \best{Bold}, and the second-best results are marked with \second{Underlined}.}
\label{tab:full_results}
\resizebox{\textwidth}{!}{
\begin{tabular}{c|c|cc|cc|cc|cc|cc|cc|cc|cc|cc}
\hline
\multirow{2}{*}{Dataset} & \multirow{2}{*}{Horizon} 
& \multicolumn{2}{c|}{\sysname} 
& \multicolumn{2}{c|}{PatchTST} 
& \multicolumn{2}{c|}{iTransformer} 
& \multicolumn{2}{c|}{Crossformer}
& \multicolumn{2}{c}{FEDformer}
& \multicolumn{2}{c|}{TimeBase} 
& \multicolumn{2}{c|}{SparseTSF} 
& \multicolumn{2}{c|}{FITS} 
& \multicolumn{2}{c}{TimeMixer} \\ 
\cline{3-20}
 & & MSE & MAE & MSE & MAE & MSE & MAE & MSE & MAE & MSE & MAE & MSE & MAE & MSE & MAE & MSE & MAE & MSE & MAE \\
\hline
\multirow{4}{*}{ETTh1} 
 & 96  & \best{0.359} & \best{0.382} & 0.377 & 0.408 & 0.389 & 0.421 & 0.408 & 0.442 & 0.485 & 0.500 & 0.365 & \second{0.387} & \second{0.362} & 0.389 & 0.380 & 0.402 & 0.410 & 0.441 \\
 & 192 & \best{0.397} & \best{0.404} & 0.413 & 0.431 & 0.424 & 0.446 & 0.472 & 0.496 & 0.481 & 0.498 & \second{0.403} & \second{0.409} & 0.404 & 0.412 & 0.415 & 0.424 & 0.448 & 0.465 \\
 & 336 & \second{0.425} & \second{0.424} & 0.436 & 0.444 & 0.456 & 0.469 & 0.480 & 0.486 & 0.522 & 0.521 & \best{0.409} & \best{0.419} & 0.435 & 0.426 & 0.449 & 0.460 & 0.475 & 0.490 \\
 & 720 & \second{0.431} & 0.450 & 0.455 & 0.475 & 0.545 & 0.532 & 0.710 & 0.616 & 0.604 & 0.575 & 0.440 & \best{0.448} & \best{0.426} & \second{0.448} & 0.433 & 0.457 & 0.475 & 0.500 \\
\hline
\multirow{4}{*}{ETTh2} 
 & 96  & \second{0.275} & \second{0.338} & 0.276 & 0.339 & 0.305 & 0.361 & 1.164 & 0.744 & 0.401 & 0.451 & 0.292 & 0.350 & 0.294 & 0.346 & \best{0.271} & \best{0.336} & 0.315 & 0.380 \\
 & 192 & 0.341 & \second{0.376} & 0.342 & 0.385 & 0.405 & 0.421 & 1.414 & 0.830 & 0.425 & 0.464 & \second{0.339} & 0.387 & 0.340 & 0.377 & \best{0.332} & \best{0.374} & 0.383 & 0.415 \\
 & 336 & 0.369 & 0.405 & 0.364 & 0.405 & 0.411 & 0.436 & 1.220 & 0.794 & 0.427 & 0.471 & 0.394 & 0.420 & \second{0.360} & \second{0.398} & \best{0.355} & \best{0.396} & 0.415 & 0.436 \\
 & 720 & 0.402 & 0.436 & 0.395 & 0.434 & 0.448 & 0.470 & 2.074 & 1.103 & 0.462 & 0.493 & 0.400 & 0.448 & \second{0.383} & \second{0.425} & \best{0.378} & \best{0.423} & 0.432 & 0.471 \\
\hline
\multirow{4}{*}{ETTm1} 
 & 96  & \best{0.293} & \best{0.344} & 0.298 & 0.352 & 0.315 & 0.369 & 0.306 & 0.353 & 0.406 & 0.441 & 0.311 & 0.351 & 0.314 & 0.359 & 0.313 & 0.357 & 0.332 & 0.384 \\
 & 192 & \best{0.323} & \best{0.361} & 0.335 & 0.373 & 0.349 & 0.388 & 0.341 & 0.385 & 0.450 & 0.477 & 0.338 & 0.371 & 0.348 & 0.376 & \second{0.339} & \second{0.369} & 0.362 & 0.398 \\
 & 336 & \best{0.358} & \best{0.381} & \second{0.366} & \second{0.389} & 0.381 & 0.409 & 0.383 & 0.420 & 0.436 & 0.466 & 0.364 & 0.386 & 0.368 & 0.386 & 0.367 & 0.385 & 0.386 & 0.413 \\
 & 720 & \best{0.412} & \best{0.410} & \second{0.420} & \second{0.421} & 0.437 & 0.439 & 0.532 & 0.512 & 0.462 & 0.479 & 0.413 & 0.414 & 0.419 & 0.413 & 0.417 & 0.417 & 0.452 & 0.457 \\
\hline
\multirow{4}{*}{ETTm2} 
 & 96  & \second{0.163} & \best{0.256} & 0.165 & 0.260 & 0.179 & 0.274 & 0.244 & 0.338 & 0.339 & 0.406 & \best{0.162} & \best{0.256} & 0.167 & 0.259 & 0.166 & \best{0.256} & 0.192 & 0.285 \\
 & 192 & \second{0.219} & \best{0.293} & \second{0.219} & \second{0.298} & 0.239 & 0.314 & 0.350 & 0.412 & 0.397 & 0.452 & \best{0.218} & \best{0.293} & \second{0.219} & 0.297 & 0.271 & 0.328 & 0.307 & 0.362 \\
 & 336 & \second{0.269} & \best{0.326} & \best{0.268} & 0.333 & 0.309 & 0.356 & 0.400 & 0.431 & 0.418 & 0.452 & 0.270 & \second{0.328} & 0.271 & 0.330 & 0.352 & 0.380 & 0.380 & 0.412 \\
 & 720 & \best{0.351} & \best{0.379} & \second{0.352} & 0.386 & 0.387 & 0.407 & 0.574 & 0.525 & 0.451 & 0.499 & \second{0.352} & \second{0.380} & 0.353 & 0.380 & \second{0.352} & \second{0.380} & 0.380 & 0.412 \\
\hline
\multirow{4}{*}{Weather} 
 & 96  & \second{0.148} & \best{0.195} & 0.149 & 0.199 & 0.159 & 0.212 & 0.151 & 0.210 & 0.289 & 0.342 & \best{0.146} & \second{0.198} & 0.174 & 0.231 & 0.176 & 0.232 & 0.163 & 0.223 \\
 & 192 & \second{0.193} & \best{0.237} & \second{0.193} & 0.243 & 0.203 & 0.252 & 0.220 & 0.273 & 0.340 & 0.403 & \best{0.185} & \second{0.241} & 0.216 & 0.267 & 0.203 & 0.256 & 0.201 & 0.255 \\
 & 336 & \second{0.242} & \best{0.278} & \best{0.240} & \second{0.281} & 0.253 & 0.291 & 0.287 & 0.342 & 0.370 & 0.408 & \second{0.263} & 0.281 & 0.260 & 0.299 & 0.261 & 0.299 & 0.258 & 0.300 \\
 & 720 & \best{0.309} & \best{0.332} & \second{0.312} & \second{0.334} & 0.317 & 0.337 & 0.362 & 0.393 & 0.420 & 0.421 & 0.314 & \second{0.331} & 0.325 & 0.345 & 0.325 & 0.346 & 0.329 & 0.348 \\
 \hline
\multirow{4}{*}{Electricity} 
 & 96  & \best{0.129} & \best{0.221} & 0.141 & 0.240 & \second{0.135} & 0.233 & 0.140 & 0.237 & 0.226 & 0.341 & 0.139 & \second{0.231} & 0.139 & 0.239 & 0.147 & 0.253 & 0.142 & 0.247 \\
 & 192 & \best{0.148} & \best{0.238} & 0.156 & 0.256 & 0.155 & 0.253 & 0.165 & 0.259 & 0.220 & 0.336 & \second{0.153} & \second{0.245} & 0.155 & 0.250 & 0.159 & 0.256 & 0.164 & 0.273 \\
 & 336 & \best{0.165} & \best{0.257} & 0.172 & 0.267 & \second{0.169} & 0.267 & 0.190 & 0.286 & 0.224 & 0.337 & \second{0.169} & 0.262 & 0.171 & 0.265 & \second{0.169} & 0.270 & 0.171 & \second{0.260} \\
 & 720 & \best{0.201} & \best{0.285} & 0.208 & 0.299 & \second{0.204} & 0.301 & 0.227 & 0.312 & 0.271 & 0.378 & 0.207 & \second{0.294} & 0.208 & 0.300 & 0.214 & 0.302 & 0.209 & 0.313 \\
 \hline
 \multirow{4}{*}{Traffic} 
 & 96  & \best{0.361} & \best{0.238} & \second{0.363} & \second{0.250} & 0.374 & 0.273 & 0.512  & 0.265  & 0.664 & 0.431 & 0.394 & 0.267 & 0.389 & 0.272 & 0.374 & 0.273 & 0.404 & 0.293 \\
 & 192 & \best{0.373} & \best{0.243} & \second{0.382} & \second{0.258} & 0.393 & 0.283 & 0.528  & 0.271  & 0.613 & 0.382 & 0.407 & 0.270 & 0.399 & 0.272 & 0.393 & 0.282 & 0.404 & 0.292 \\
 & 336 & \best{0.385} & \best{0.248} & \second{0.399} & \second{0.268} & 0.409 & 0.292 & 0.543  & 0.281  & 0.612 & 0.379 & 0.417 & 0.278 & 0.417 & 0.279 & 0.423 & 0.292 & 0.425 & 0.293 \\
 & 720 & \best{0.428} & \best{0.270} & \second{0.432} & \second{0.289} & 0.450 & 0.314 & 0.598  & 0.314  & 0.664 & 0.410 & 0.456 & 0.298 & 0.449 & 0.299 & 0.450 & 0.314 & 0.453 & 0.314 \\
 \hline
\end{tabular}}
\end{table}

\begin{table}[t]
\centering
\caption{Impact of router number $R$ on prediction accuracy. Each entry reports MSE and MAE. The input length is set to 720. The best results are marked with \best{Bold}, and the second-best results are marked with \second{Underlined}.}
\label{tab:router_number}
\small
\begin{tabular}{c|cc|cc|cc|cc|cc}
\hline
\multirow{2}{*}{Dataset} 
& \multicolumn{2}{c|}{R=1} 
& \multicolumn{2}{c|}{R=2} 
& \multicolumn{2}{c|}{R=4} 
& \multicolumn{2}{c|}{R=8} 
& \multicolumn{2}{c}{R=16} \\
\cline{2-11}
 & MSE & MAE 
 & MSE & MAE 
 & MSE & MAE 
 & MSE & MAE 
 & MSE & MAE \\
\hline
Weather 
& 0.162 & 0.210
& 0.153 & 0.199
& 0.151 & 0.199
& \best{0.148} & \best{0.195}
& \second{0.149} & \second{0.198} \\
Traffic 
& 0.372 & 0.249 
& \second{0.367} & 0.243 
& \best{0.361} & \best{0.238} 
& 0.364 & \second{0.242} 
& 0.368 & 0.243 \\
Electricity 
& 0.133 & 0.228 
& \second{0.132} & \second{0.226} 
& \best{0.129} & \best{0.221} 
& 0.130 & 0.223 
& 0.130 & 0.222 \\
\hline
\end{tabular}
\end{table}

\subsection{Additional Hyper Parameters Analysis}
\subsubsection{Impact of Model Parameter Scale on Performance}
We conducted comparative experiments on three variants of the \sysname model with different parameter scales across the Electricity, Traffic, and Weather datasets. The three configurations are: a single-layer model with latent dimension 8 ($\approx$1.72K parameters), a single-layer model with latent dimension 16 ($\approx$5.48K parameters), and a two-layer model with latent dimension 32 ($\approx$37.1K parameters). \figref{tab:phaseformer_params_variants} summarizes the results in terms of MSE, MAE and FLOPs.

Overall, the effect of model scale on performance is not consistent. On the Traffic dataset, larger models yield slight improvements, whereas on the Electricity and Weather datasets, the smaller and medium-sized models perform better. These findings indicate that \sysname achieves a favorable balance between computational efficiency and predictive accuracy at relatively small parameter scales, and increasing model size does not lead to uniform gains across all tasks.

\begin{table}[t]
\centering
\caption{Comparison of \sysname variants. Each cell reports MSE, MAE, and FLOPs. The input length is fixed as 720 steps and the output length is fixed as 96 steps. The best results are marked with \best{Bold}, and the second-best results are marked with \second{Underlined}.}
\label{tab:phaseformer_params_variants}
\small  
\begin{tabular}{c|ccc|ccc|ccc}
\hline
\multirow{2}{*}{Dataset}
& \multicolumn{3}{c|}{\textbf{\sysname-1.7K}} 
& \multicolumn{3}{c|}{\textbf{\sysname-5K}} 
& \multicolumn{3}{c}{\textbf{\sysname-37K}} \\
\cline{2-10}
& MSE & MAE & FLOPs
& MSE & MAE & FLOPs
& MSE & MAE & FLOPs \\
\hline
Electricity 
& \best{0.129} & \best{0.220} & \best{9.41M}
& \best{0.129} & \second{0.221} & \second{31.97M}
& \second{0.131} & \second{0.223} & 221.05M \\
Traffic     
& \second{0.361} & \second{0.241} & \best{25.27M}
& 0.366 & 0.243 & \second{85.84M}
& \best{0.360} & \best{0.236} & 593.61M \\
Weather     
& \best{0.150} & \second{0.199} & \best{0.62M}
& \second{0.151} & \best{0.194} & \second{2.09M}
& 0.174 & 0.217 & 14.46M \\
\hline
\end{tabular}
\end{table}

\subsubsection{Impact of Input Length on Performance}

\begin{figure}[htbp]
    \centering
    \begin{subfigure}[b]{0.32\textwidth}
        \centering
        \includegraphics[width=\textwidth]{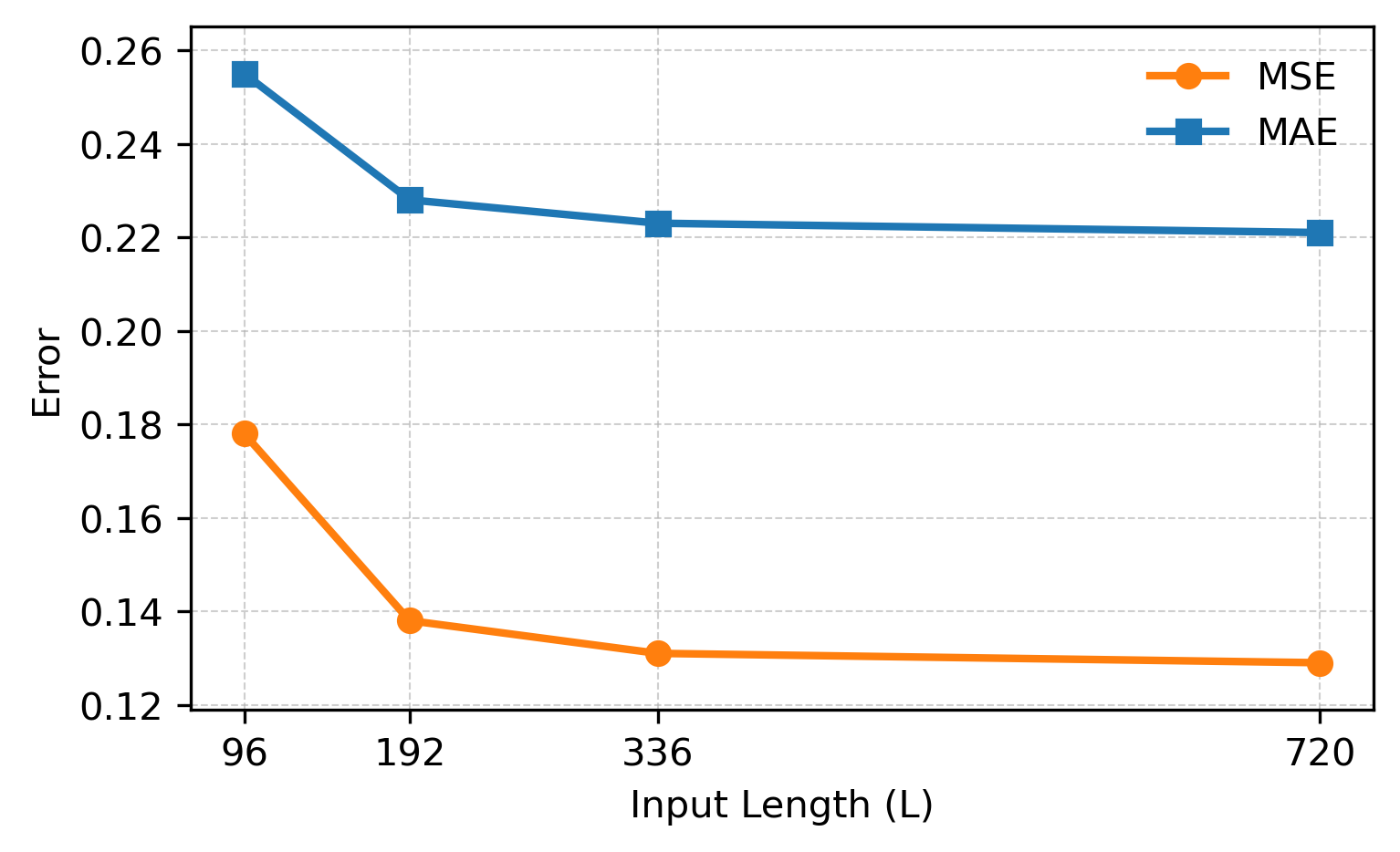}
        \caption{Electricity}
    \end{subfigure}
    \begin{subfigure}[b]{0.32\textwidth}
        \centering
        \includegraphics[width=\textwidth]{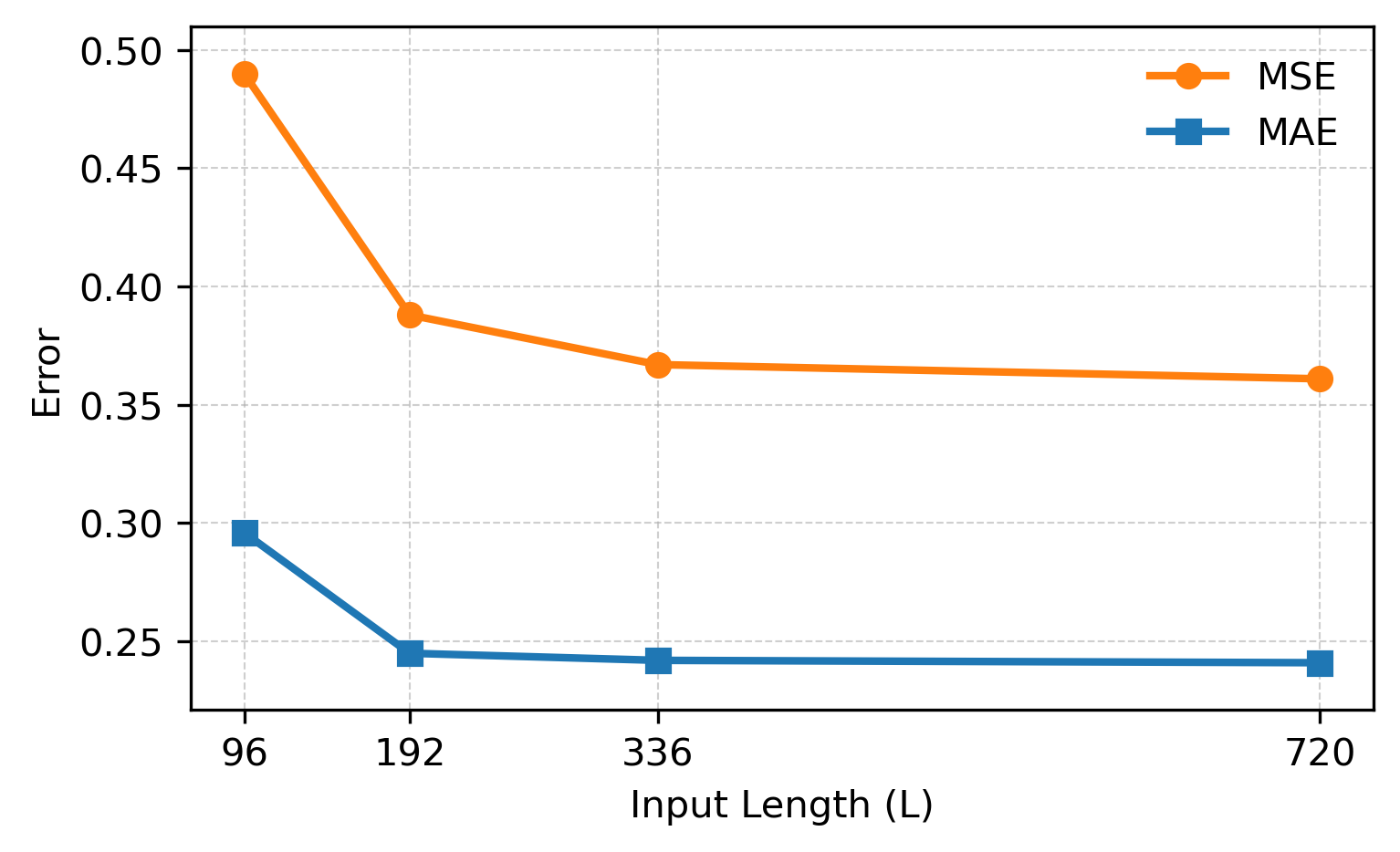}
        \caption{Traffic}
    \end{subfigure}
    \begin{subfigure}[b]{0.32\textwidth}
        \centering
        \includegraphics[width=\textwidth]{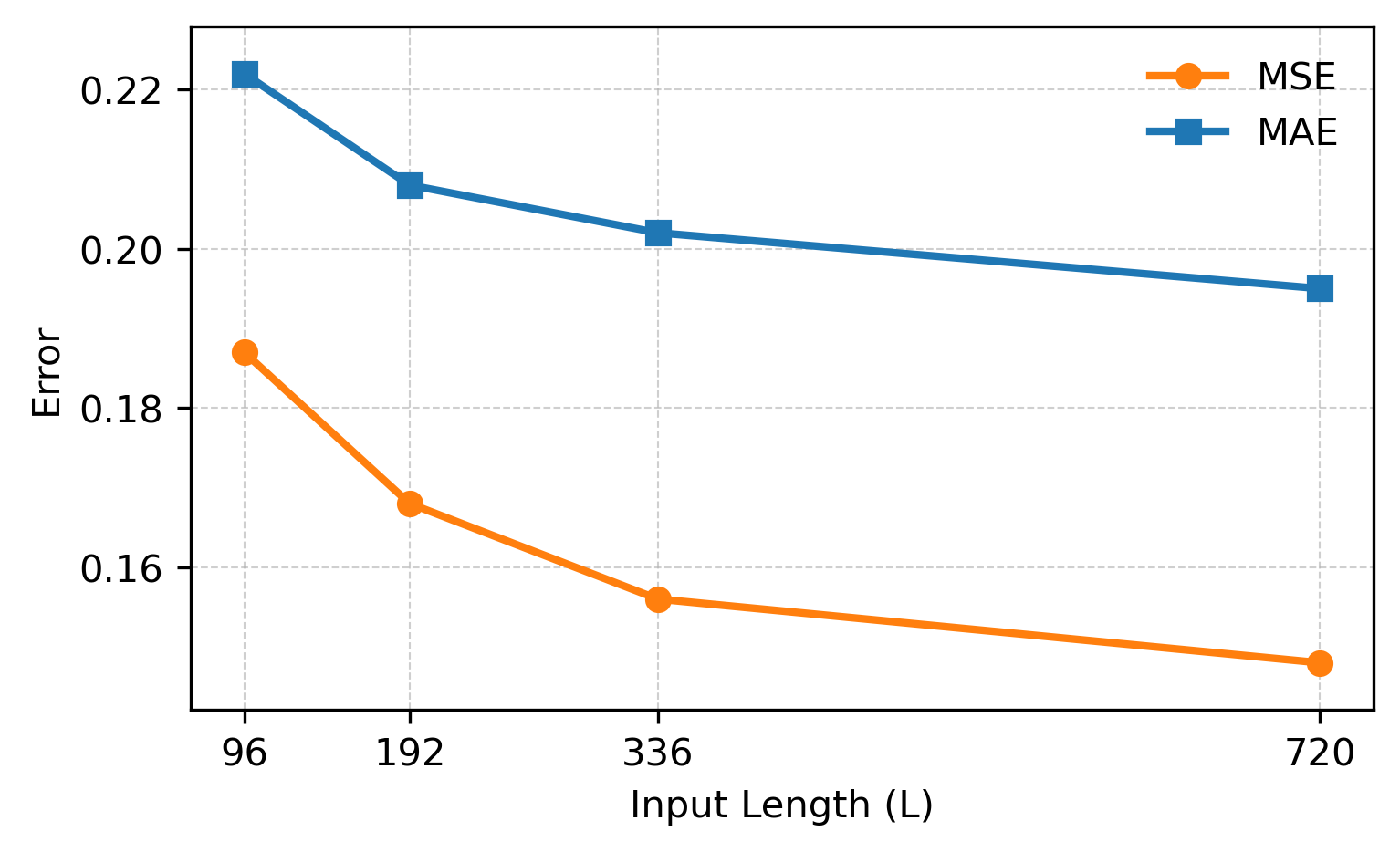}
        \caption{Weather}
    \end{subfigure}

    \caption{Prediction error test results across datasets under different input lengths.}
    \label{fig:input_length_prediction_errors}
\end{figure}

We examine how the input window size affects the prediction accuracy and computational cost of \sysname. 
Throughout this section, $L$ denotes using the most recent $L$ time steps as model input. The output length is fixed as 96 steps.
As summarized in ~\figref{fig:input_length_prediction_errors}, increasing $L$ reduces MSE and MAE across datasets, indicating that \sysname benefits from longer historical context for modeling long-range temporal dependencies.

\begin{figure}[htbp]
    \centering
    \begin{subfigure}[b]{0.32\textwidth}
        \centering
        \includegraphics[width=\textwidth]{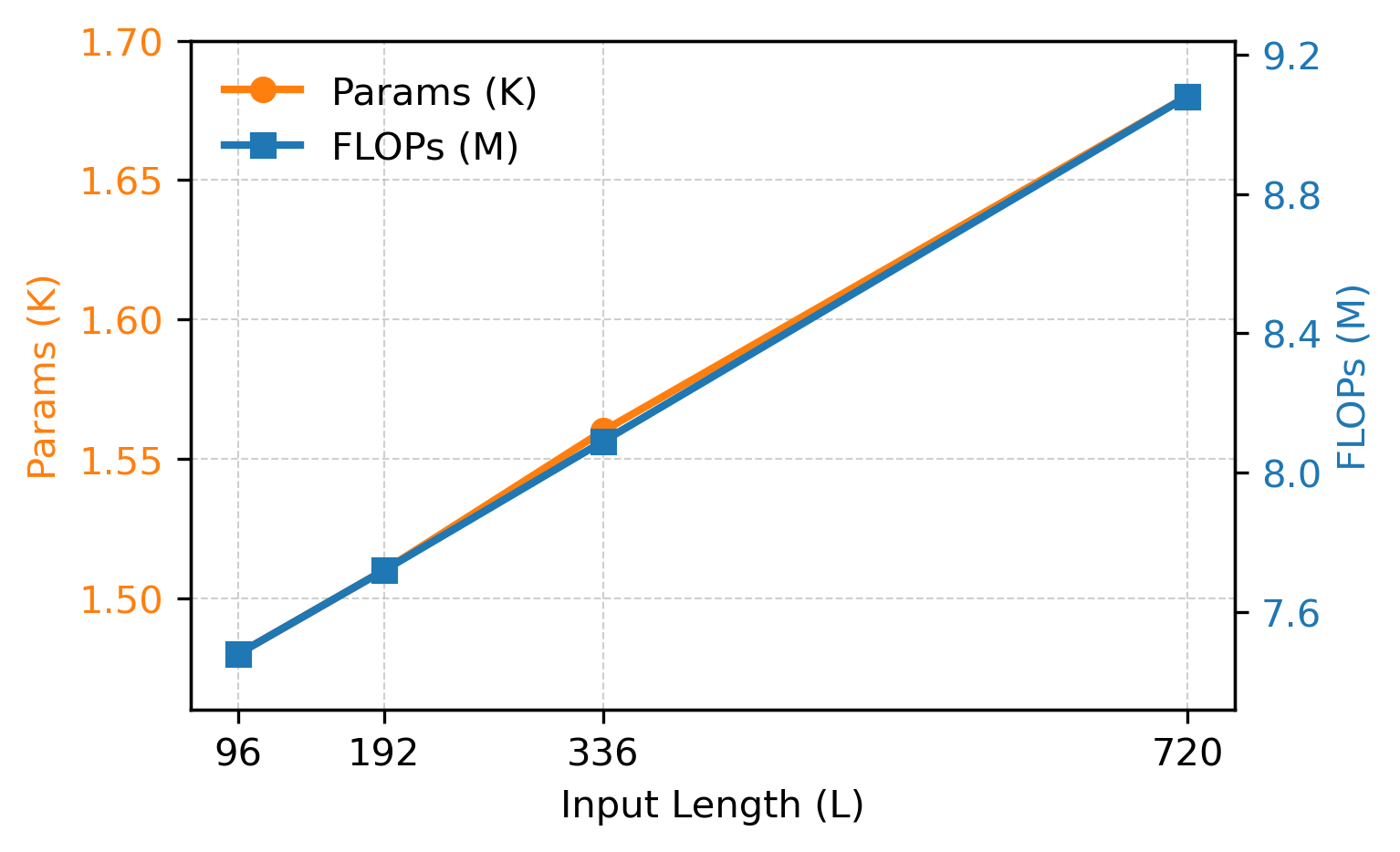}
        \caption{Electricity}
    \end{subfigure}
    \begin{subfigure}[b]{0.32\textwidth}
        \centering
        \includegraphics[width=\textwidth]{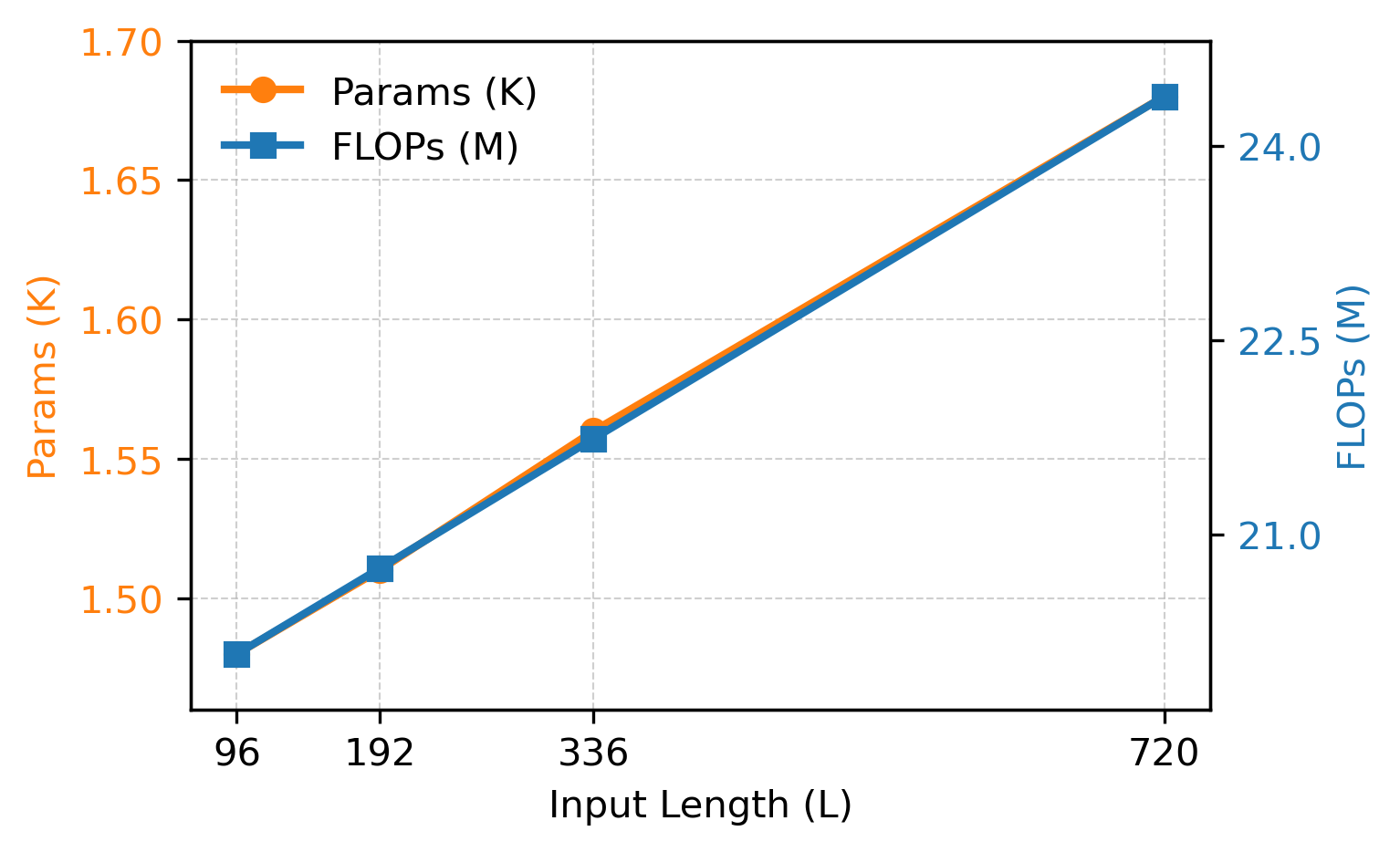}
        \caption{Traffic}
    \end{subfigure}
    \begin{subfigure}[b]{0.32\textwidth}
        \centering
        \includegraphics[width=\textwidth]{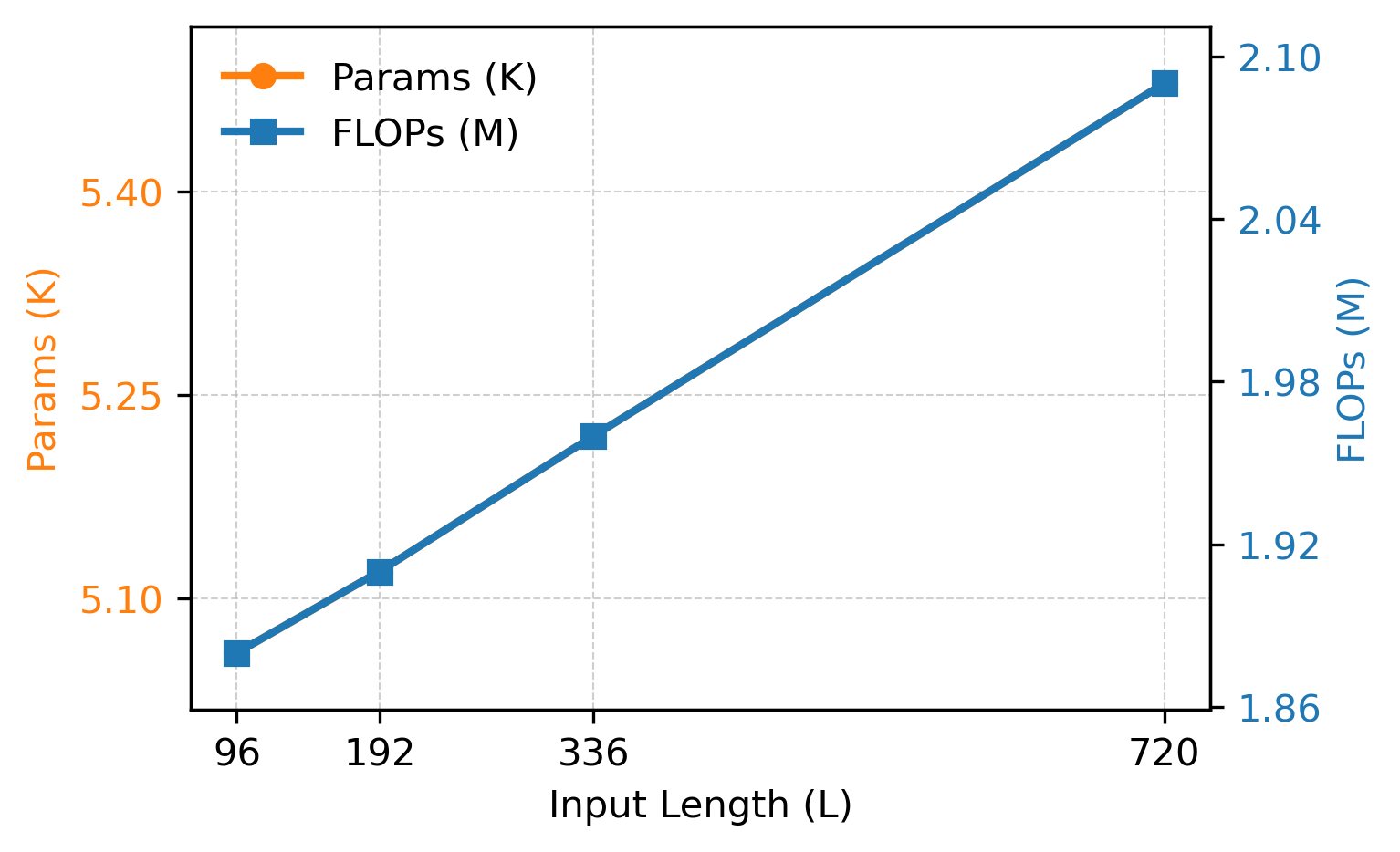}
        \caption{Weather}
    \end{subfigure}

    \caption{Efficiency evaluation of \sysname\ across datasets with varying input lengths.}
    \label{fig:input_length_efficiency}
\end{figure}

In terms of efficiency, the parameter count and FLOPs per forward pass remain nearly constant as $L$ increases, with only modest growth (see ~\figref{fig:input_length_efficiency}) attributable primarily to the embedding stage. 
This behavior arises because the sequence length processed by the core encoder/decoder is governed by the number of \emph{phases}, which depends on the data’s learned periodic structure rather than by the raw input length.
Consequently, scaling $L$ mainly affects the embedding computations, whose cost is relatively small compared to the phase-based modules.

\subsection{Showcases}

To provide a clearer comparison of predictive performance across different models, we present the results of \sysname, PatchTST, FITS, and TimeBase on the Traffic dataset.
\sysname demonstrates strong predictive performance, as reflected by both the shape of its forecasts and the actual prediction errors.

Compared with \sysname, PatchTST produces lower peak values within each cycle, failing to fully match the true curve. This discrepancy is likely due to phase shifts in the periodic pattern that reduce peak amplitudes.
FITS, which performs prediction in the frequency domain with frequency band partitioning, tends to overlook high-frequency information. As a result, its predictions deviate from the ground truth, though the forecasts still preserve a periodic structure.
TimeBase aligns well with the general cyclical pattern but fails to capture true variations across cycles, a limitation stemming from its patch-based basis construction mechanism.

\begin{figure}[t]
    \centering
    \begin{subfigure}{0.47\textwidth}
        \centering
        \includegraphics[width=\linewidth]{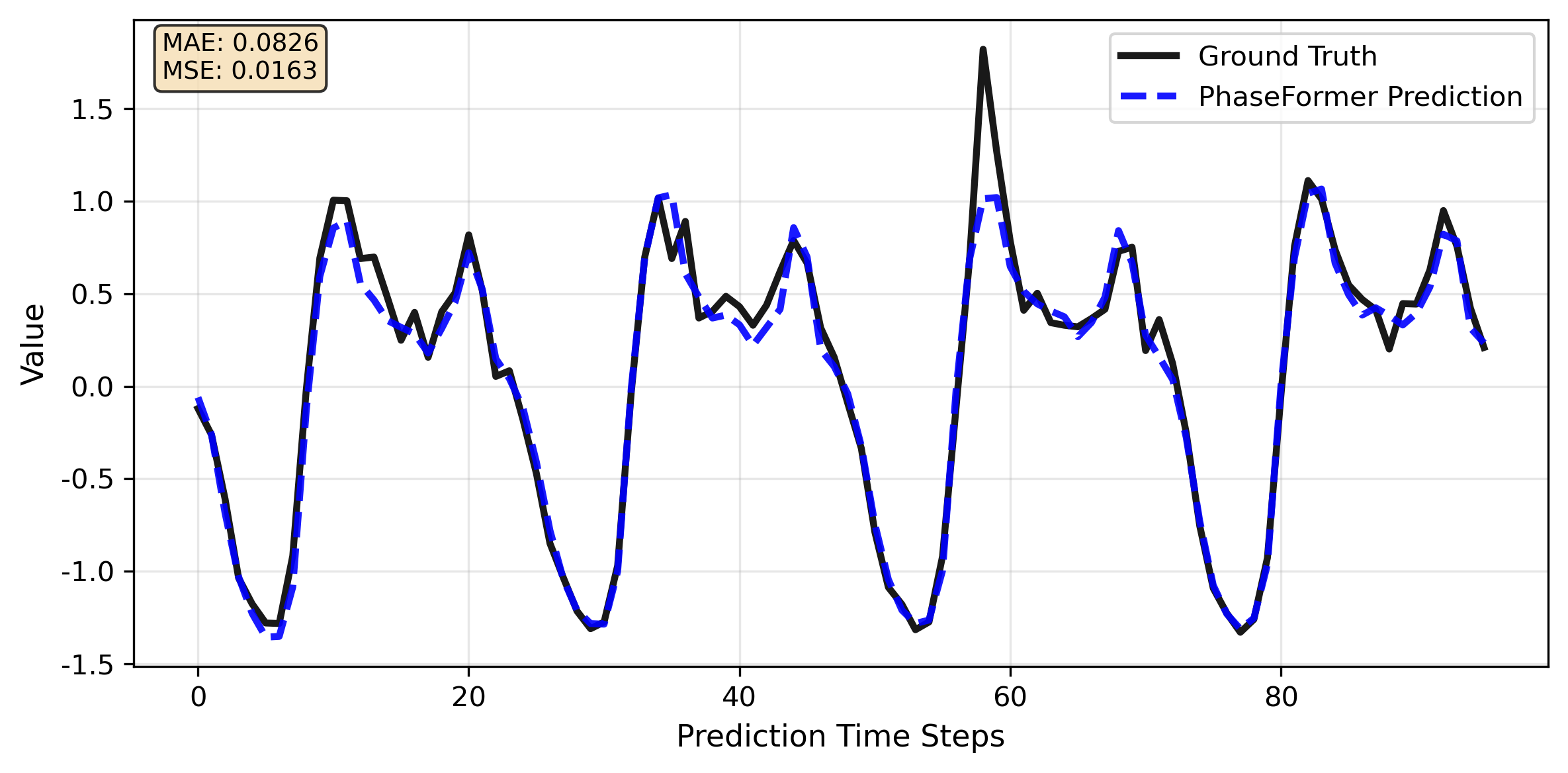}
        \caption{\sysname}
    \end{subfigure}
    \hfill
    \begin{subfigure}{0.47\textwidth}
        \centering
        \includegraphics[width=\linewidth]{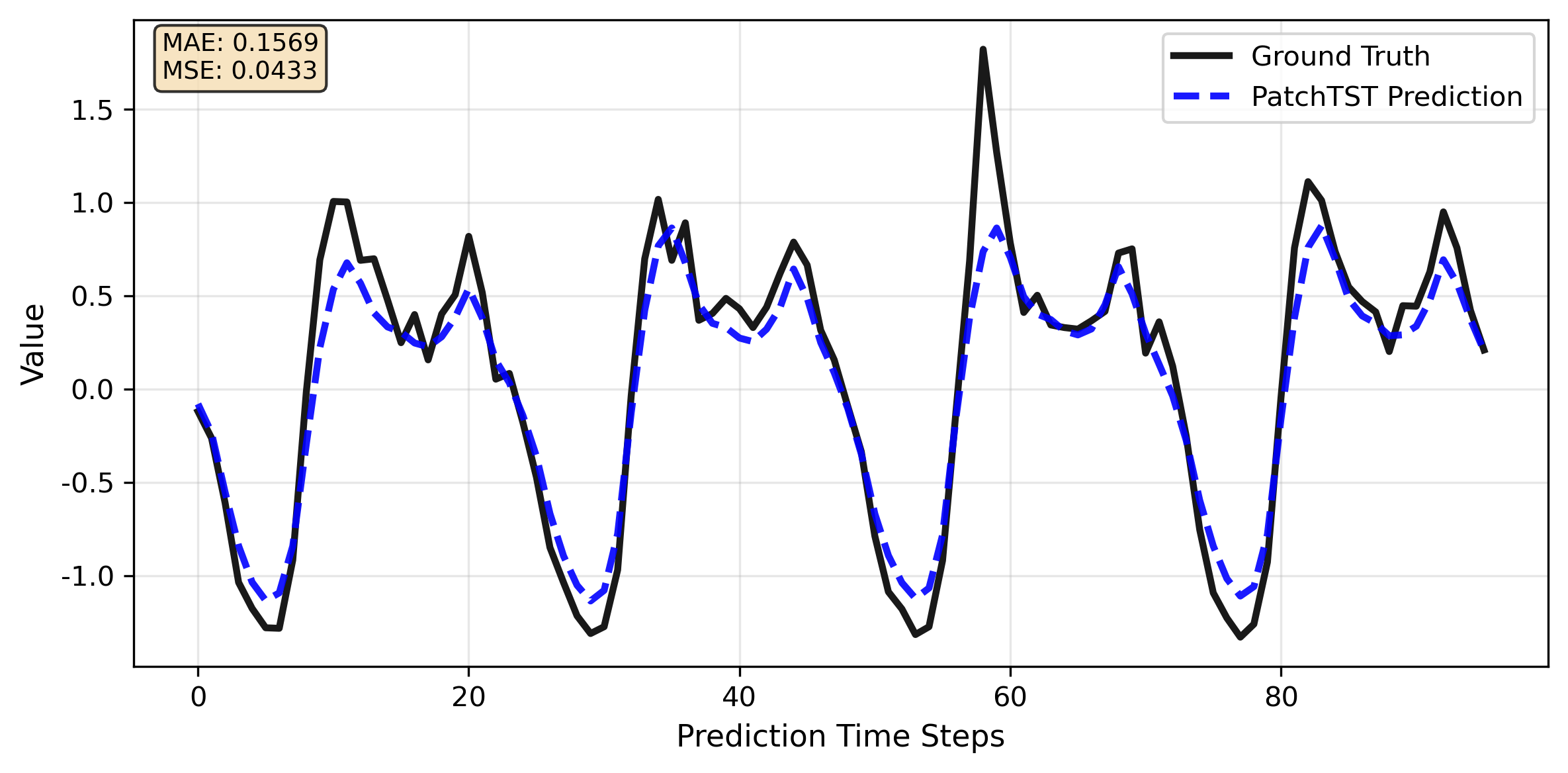}
        \caption{PatchTST}
    \end{subfigure}

    \vskip\baselineskip
    \begin{subfigure}{0.47\textwidth}
        \centering
        \includegraphics[width=\linewidth]{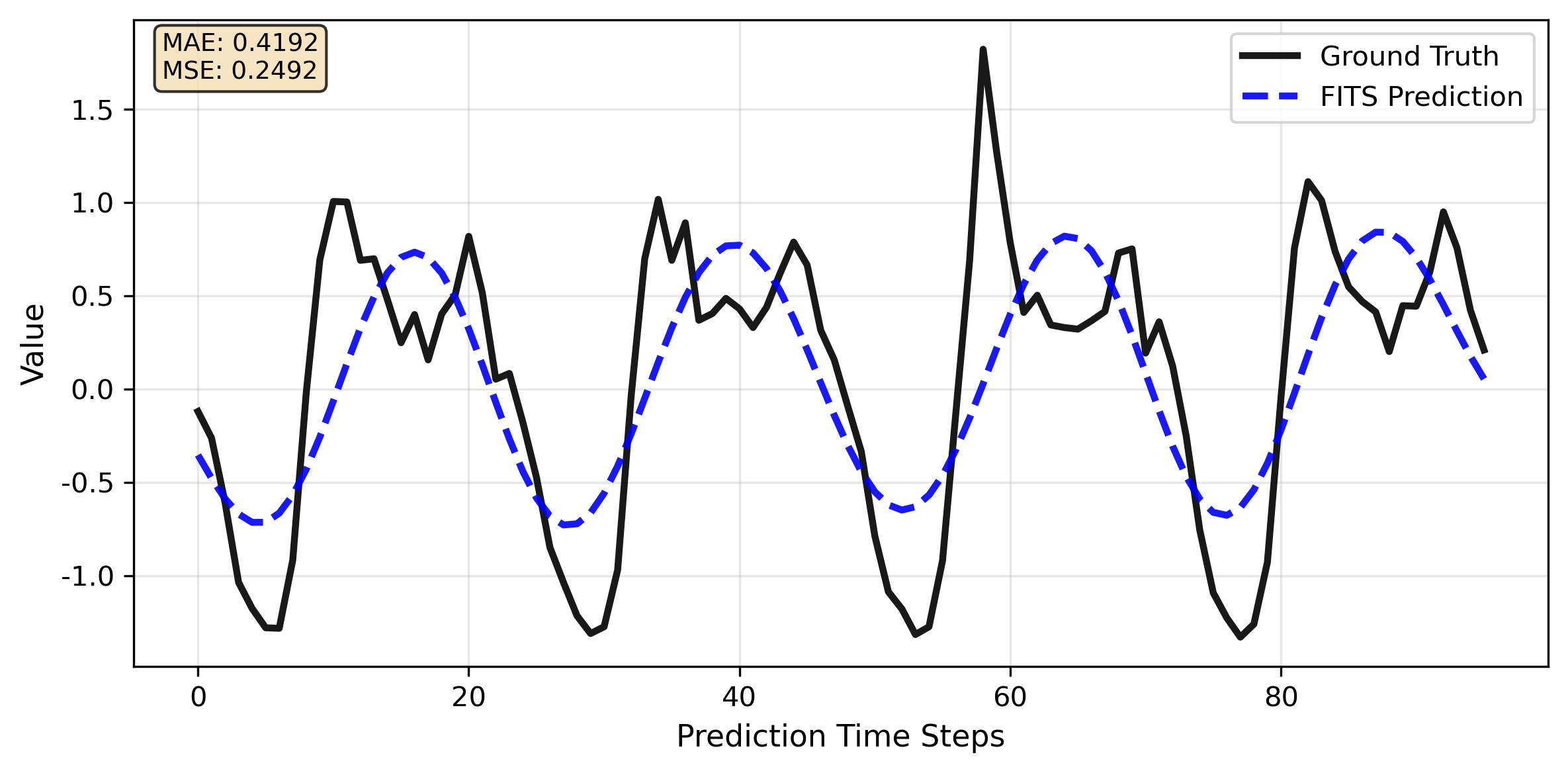}
        \caption{FITS}
    \end{subfigure}
    \hfill
    \begin{subfigure}{0.47\textwidth}
        \centering
        \includegraphics[width=\linewidth]{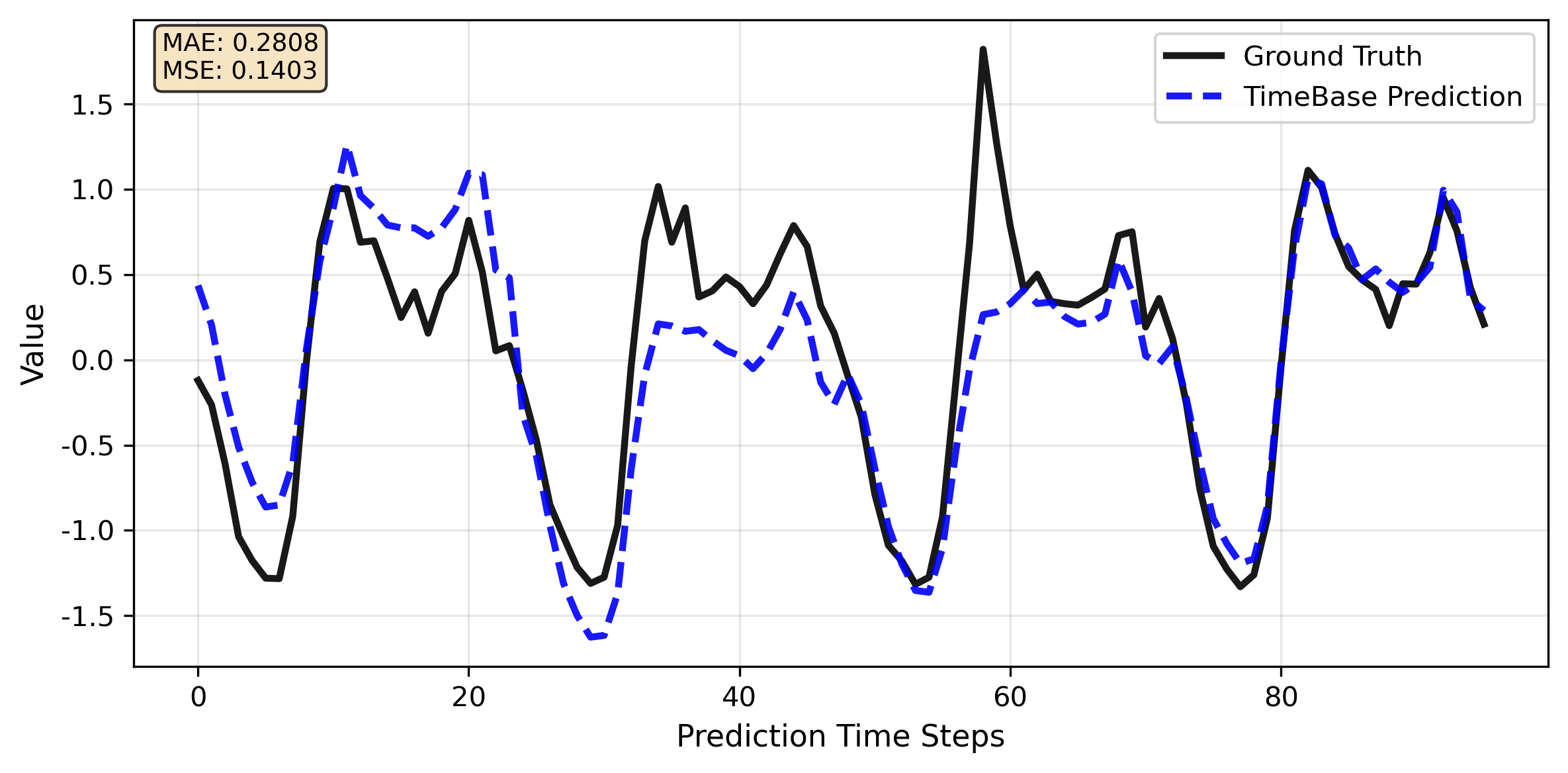}
        \caption{TimeBase}
    \end{subfigure}

    \caption{Visualization of forecasting results on Traffic dataset. The black lines stand for the ground truth and the \textcolor{blue}{blue} lines stand for forecasting results.}
\end{figure}

\subsection{Theoretical Analysis for Phase Tokenization}
\label{sub:theoretical}
We model the data matrix as
\[
X \;=\; A G^\top + N \;\in\;\mathbb{R}^{D\times H},
\]
where $A\in\mathbb{R}^{D\times r}$, $G\in\mathbb{R}^{H\times r}$ are column full rank with $\operatorname{rank}(A)=\operatorname{rank}(G)=r$, and $N$ is noise. The true signal is $M = A G^\top$. We assume $r\ll \min(D,H)$.

Patch tokenization corresponds to the row space $\mathrm{Row}(X)$ (the right singular $r$-subspace), while phase tokenization corresponds to the column space $\mathrm{Col}(X)$ (the left singular $r$-subspace).

A shared transformation applies $S\in\mathbb{R}^{H\times H}$ on the hourly dimension:
\[
X’ = X S^\top = A (S G)^\top + N’.
\]

For a matrix $Y$, define the spectral separation as
\[
\mathrm{sep}_r(Y) := \min_{i\le r,\,j>r}|\sigma_i(Y)-\sigma_j(Y)|.
\]
When $\operatorname{rank}(Y)=r$, this equals $\sigma_r(Y)$. In particular,
\[
\delta = \sigma_r(M), \quad \delta'=\sigma_r(MS^\top), \quad
\delta_{\min}=\min(\delta,\delta').
\]
Assume $S$ is invertible on $\mathrm{Col}(G)$, i.e. $\operatorname{rank}(S G)=r$. Define
\[
\kappa := \sigma_{\min}(S|_{\mathrm{Col}(G)})>0.
\]
Moreover, since $\sigma_r(M)\ge\sigma_r(A)\sigma_r(G)$ and 
$\sigma_r(MS^\top)\ge \kappa\sigma_r(A)\sigma_r(G)$, we have the useful bound
\[
\delta_{\min}\;\ge\;\min(1,\kappa)\,\sigma_r(A)\sigma_r(G).
\]

For two $r$-dimensional subspaces $\mathcal U,\mathcal V$, their distance is
\[
d(\mathcal U,\mathcal V) := \|P_{\mathcal U}-P_{\mathcal V}\|_2 \;=\; \sin\Theta_{\max}(\mathcal U,\mathcal V),
\]
where $P_{\mathcal U}$ is the orthogonal projector onto $\mathcal U$. 
This metric satisfies the triangle inequality.

\begin{lemma}[Column space preservation]
Let $M=A G^\top$. If $\operatorname{rank}(S G)=r$, then
\[
\mathrm{Col}(M S^\top) = \mathrm{Col}(M) = \mathrm{Col}(A).
\]
If $\operatorname{rank}(S G)<r$, then the column space shrinks.
\end{lemma}

\begin{lemma}[Row space change]
For $M=A G^\top$,
\[
\mathrm{Row}(M) = \mathrm{Col}(G), \qquad \mathrm{Row}(M S^\top) = \mathrm{Col}(S G).
\]
As a result:
\[
d(\mathrm{Col}(G),\mathrm{Col}(S G))>0 \quad \Longleftrightarrow \quad S(\mathrm{Col}(G)) \neq \mathrm{Col}(G).
\]
\end{lemma}

\begin{lemma}[Wedin’s $\sin\Theta$ theorem]
Let $\widehat M=M+E$ and $\delta=\mathrm{sep}_r(M)>0$. Then
\[
d\big(\mathcal U_r(M),\mathcal U_r(\widehat M)\big) \;\le\; C \frac{\|E\|_2}{\delta},
\]
where $\mathcal U_r(M)$ denotes the leading left singular $r$-subspace of $M$ (the right case is analogous). 
Here $C$ is an absolute constant (often $C\in[2,2\sqrt2]$). The condition $\delta>0$ is necessary.
\end{lemma}

\begin{theorem}
Assume $\operatorname{rank}(S G)=r$ and $\delta,\delta’>0$. Then:

\begin{enumerate}
\item For phase tokenization,
\[
d\big(\mathcal U_r(X), \mathcal U_r(X’)\big) \;\le\;
C\Big(\tfrac{\|N\|_2}{\delta}+\tfrac{\|N’\|_2}{\delta’}\Big)
\;\le\; C\frac{\|N\|_2+\|N’\|_2}{\delta_{\min}}.
\]
In the noiseless case, Lemma~1 ensures exact invariance, so the distance is $0$.

\item For patch tokenization,
\[
d\big(\mathcal V_r(X), \mathcal V_r(X’)\big) \;\ge\;
d_0 - C\Big(\tfrac{\|N\|_2}{\delta}+\tfrac{\|N’\|_2}{\delta’}\Big),
\]
where $d_0 := d(\mathrm{Col}(G),\mathrm{Col}(S G))$. If $S(\mathrm{Col}(G))\neq\mathrm{Col}(G)$, then $d_0>0$.
\end{enumerate}
\end{theorem}

\begin{proof}
For phase tokenization, apply the triangle inequality:
\[
d(\mathcal U_r(X),\mathcal U_r(X'))
\le d(\mathcal U_r(X),\mathcal U_r(M))
+ d(\mathcal U_r(M),\mathcal U_r(MS^\top))
+ d(\mathcal U_r(MS^\top),\mathcal U_r(X')).
\]
By Lemma~1 the middle term vanishes, and the two boundary terms are bounded by Wedin’s theorem, yielding the stated inequality.  

For patch tokenization, we have
\[
d(\mathcal V_r(X),\mathcal V_r(X'))
\ge d(\mathcal V_r(M),\mathcal V_r(MS^\top))
- d(\mathcal V_r(M),\mathcal V_r(X))
- d(\mathcal V_r(MS^\top),\mathcal V_r(X')).
\]
By Lemma~2 the first term equals $d_0$, and the other two are controlled by Wedin’s theorem, proving the bound.
\end{proof}

In real-world scenarios, slight variations in timing or conditions occur from day to day, so the daily transformations are not exactly identical. 
We model this systematic inconsistency by introducing a small perturbation $\Delta_d$, which captures the mismatch between the ideal linear transformation $S$ and the actual data-generating process.
Suppose each day’s transform is $S_d=S+\Delta_d$ with $\|\Delta_d\|_2\le\varepsilon$. Then
\[
X’ = X S^\top + R, \qquad R_{d,:}=X_{d,:}\Delta_d^\top.
\]
Bounding row by row gives $\|R_{d,:}\|_2 \le \varepsilon\|X_{d,:}\|_2$, hence
\[
\|R\|_F \le \varepsilon \|X\|_F \quad \Rightarrow \quad 
\|R\|_2 \le \varepsilon(\|M\|_F+\|N\|_F).
\]

\begin{theorem}[Stability under day-wise perturbations]\label{thm:relaxed}
Under the relaxed model, each day uses $S_d=S+\Delta_d$ with $\|\Delta_d\|_2\le\varepsilon$, so that
\[
X' \;=\; X S^\top + R,\qquad R_{d,:}=X_{d,:}\Delta_d^\top .
\]
Let $X=M+N$ with $M=AG^\top$, $\operatorname{rank}(A)=\operatorname{rank}(G)=r$, and assume
$\operatorname{rank}(SG)=r$ so that $\delta=\sigma_r(M)>0$ and $\delta'=\sigma_r(MS^\top)>0$.
Define $\delta_{\min}=\min(\delta,\delta')$ and $d_0=d(\mathrm{Col}(G),\mathrm{Col}(SG))$.
Then there exists an absolute constant $C\in[2,2\sqrt2]$ such that:
\begin{enumerate}[leftmargin=2em]
\item \textbf{Phase tokenization (left $r$-subspace):}
\[
\begin{aligned}
d\big(\mathcal U_r(X), \mathcal U_r(X')\big)
&\;\le\;
C\Bigg(
   \frac{\|N\|_2}{\delta}
 + \frac{\|N'\|_2}{\delta'}
 + \frac{\|R\|_2}{\delta'}
\Bigg) \\
&\;\le\;
C\,\frac{\varepsilon(\|M\|_F+\|N\|_F)
   + \|N\|_2
   + \|N'\|_2}{\delta_{\min}}.
\end{aligned}
\]

\item \textbf{Patch tokenization (right $r$-subspace):}
\[
\begin{aligned}
d\big(\mathcal V_r(X), \mathcal V_r(X')\big)
&\;\ge\;
d_0 \;-\;
C\Bigg(
   \frac{\|N\|_2}{\delta}
 + \frac{\|N'\|_2}{\delta'}
 + \frac{\|R\|_2}{\delta'}
\Bigg) \\
&\;\ge\;
d_0 \;-\;
C\,\frac{\varepsilon(\|M\|_F+\|N\|_F)
   + \|N\|_2
   + \|N'\|_2}{\delta_{\min}}.
\end{aligned}
\]
In particular, if $S(\mathrm{Col}(G))=\mathrm{Col}(G)$ then $d_0=0$ and patch tokenization is also preserved up to the same perturbation scale.
\end{enumerate}
Moreover, using $\delta_{\min}\ge \min(1,\kappa)\,\sigma_r(A)\sigma_r(G)$ with
$\kappa=\sigma_{\min}(S|_{\mathrm{Col}(G)})>0$ makes the role of signal strength explicit.
\end{theorem}

\begin{proof}
By row-wise control,
$\|R_{d,:}\|_2\le \varepsilon\|X_{d,:}\|_2$, hence
\[
\|R\|_F\le \varepsilon\|X\|_F\le \varepsilon(\|M\|_F+\|N\|_F),
\qquad
\|R\|_2\le \|R\|_F\le \varepsilon(\|M\|_F+\|N\|_F).
\]

Insert the chain
\[
X \;\to\; M \;\to\; MS^\top \;\to\; MS^\top+N' \;\to\; X'=MS^\top+N'+R.
\]

For phase subspace $\mathcal U_r$, according to the triangle inequality,
\[
\begin{aligned}
d\big(\mathcal U_r(X),\mathcal U_r(X')\big)
&\le d\big(\mathcal U_r(X),\mathcal U_r(M)\big)
+ d\big(\mathcal U_r(M),\mathcal U_r(MS^\top)\big)\\
&\quad + d\big(\mathcal U_r(MS^\top),\mathcal U_r(MS^\top+N')\big)
+ d\big(\mathcal U_r(MS^\top+N'),\mathcal U_r(X')\big).
\end{aligned}
\]
The middle term vanishes by Column space preservation (Lemma~1).
Applying Wedin’s $\sin\Theta$ theorem (Lemma~3) to the remaining three perturbations $E\in\{N,\;N',\;R\}$
yields
$C\|N\|_2/\delta + C\|N'\|_2/\delta' + C\|R\|_2/\delta'$.
Use $\delta_{\min}\le\delta,\delta'$ and Step~1 to obtain Item~1.

For patch subspace $\mathcal V_r$, we use the reverse triangle inequality:
\[
\begin{aligned}
d\big(\mathcal V_r(X),\mathcal V_r(X')\big)
&\ge d\big(\mathcal V_r(M),\mathcal V_r(MS^\top)\big)
- d\big(\mathcal V_r(M),\mathcal V_r(X)\big)\\
&\quad - d\big(\mathcal V_r(MS^\top),\mathcal V_r(MS^\top+N')\big)
- d\big(\mathcal V_r(MS^\top+N'),\mathcal V_r(X')\big).
\end{aligned}
\]
The first term equals $d_0$ by Row space change (Lemma~2).
Apply Wedin’s theorem to the other three terms to obtain the final results.
\end{proof}

\subsection{The Use of Large Language Models}

In this work, large language models (specifically ChatGPT-5) are used solely for polishing the writing, identifying grammatical issues, and performing proofreading.

\end{document}